\documentclass[letterpaper]{article} 
\usepackage{aaai24}  
\usepackage{times}  
\usepackage{helvet}  
\usepackage{courier}  
\usepackage[hyphens]{url}  
\usepackage{graphicx} 
\usepackage{natbib}  
\usepackage{caption} 
\usepackage{algorithm}
\usepackage{algorithmic}

\usepackage{newfloat}
\usepackage{listings}

\usepackage{multirow}
\usepackage{booktabs}
\usepackage{graphicx}
\usepackage{textcomp}
\usepackage{mathtools, nccmath}
\usepackage{lscape}

\DeclareCaptionStyle{ruled}{labelfont=normalfont,labelsep=colon,strut=off} 
\floatstyle{ruled}
\newfloat{listing}{tb}{lst}{}
\floatname{listing}{Listing}
\usepackage{amsmath}
\usepackage{amsfonts}
\usepackage{amssymb}
\usepackage{amsthm}
\newtheorem{definition}{Definition}
\newtheorem{proposition}[definition]{Proposition}
\newtheorem{corollary}[definition]{Corollary}
\newtheorem{lemma}[definition]{Lemma}
\newtheorem{theorem}[definition]{Theorem}

\title{Weisfeiler and Lehman Go Paths: Learning Topological Features via Path Complexes}
\author{
    Quang Truong\thanks{Corresponding author}, Peter Chin
}
\affiliations{
    Thayer School of Engineering, Dartmouth College \\
    cong.minh.quang.truong.th@dartmouth.edu, peter.chin@dartmouth.edu
}

\begin{document}
\nocopyright
\maketitle

\begin{abstract}
Graph Neural Networks (GNNs), despite achieving remarkable performance across different tasks, are theoretically bounded by the 1-Weisfeiler-Lehman test, resulting in limitations in terms of graph expressivity. Even though prior works on topological higher-order GNNs overcome that boundary, these models often depend on assumptions about sub-structures of graphs. Specifically, topological GNNs leverage the prevalence of cliques, cycles, and rings to enhance the message-passing procedure. Our study presents a novel perspective by focusing on simple paths within graphs during the topological message-passing process, thus liberating the model from restrictive inductive biases. We prove that by lifting graphs to path complexes, our model can generalize the existing works on topology while inheriting several theoretical results on simplicial complexes and regular cell complexes. Without making prior assumptions about graph sub-structures, our method outperforms earlier works in other topological domains and achieves state-of-the-art results on various benchmarks.
\end{abstract}

\section{Introduction} \label{section:intro}
Graph-based learning presents an intricate problem due to the inherent ambiguity in defining geometric properties such as canonical vertex ordering \cite{bouritsas_improving_2021}. Exploiting these properties is however challenging, given the pervasive utility of graphs across an array of domains. Some early prominent efforts on graph neural networks (GNNs) shed light upon this field \cite{bruna_spectral_2014, defferrard_convolutional_2016, kipf_semi-supervised_2017}, which effectively paved the path for the recent advancements \cite{xu_representation_2018, xu_how_2019, velickovic_graph_2018}. However, the graph expressivity of GNNs is upper-bounded by the 1-WL test \cite{weisfeiler_reduction_1968}, which is proven by \cite{xu_how_2019, morris_weisfeiler_2019}. This limitation has led to interests in higher-order GNNs, a sub-family of GNNs that extend beyond simple pairwise vertex interactions to encompass broader relationships \cite{huang_unignn_2021, feng_hypergraph_2018, yadati_hypergcn_2019, bodnar_weisfeiler_2021, bodnar_weisfeiler_2022, ebli_simplicial_2020, roddenberry_principled_2021, giusti_cell_2022, giusti_cin_2023, hajij_topological_2023, papillon_architectures_2023}. For example, SIN\cite{bodnar_weisfeiler_2021} and CIN \cite{bodnar_weisfeiler_2022} are proven to be not less powerful than the 3-WL test, \(k\)-GNNs \cite{morris_weisfeiler_2019} and PPGNs \cite{maron_provably_2019} are proven to be as powerful as the \(k\)-WL test \cite{grohe_pebble_2012, grohe_descriptive_2017}, and \(\delta\)-\(k\)-LGNN \cite{morris_weisfeiler_2020} is strictly more powerful than the \(k\)-WL test. Topological GNNs, which extend graphs to topological domains such as simplicial complexes \cite{bodnar_weisfeiler_2021, ebli_simplicial_2020, roddenberry_principled_2021, schaub_random_2020} and regular cell complexes \cite{bodnar_weisfeiler_2022, giusti_cell_2022, giusti_cin_2023} to learn higher-order features, demonstrate outstanding performance on graph classification tasks. Yet, these models are predicated on the premise that graphs should contain cliques, cycles, or induced cycles (rings).

Drawing inspiration from path complexes  \cite{grigoryan_path_2020, grigoryan_homologies_2013}, we relax the above assumption by focusing only on simple paths, foundational yet universal elements in graphs, during message propagation. Specifically, we lift our graphs to a topological domain referred to as \textbf{path complexes}, with \textbf{elementary paths} serving as the basis elements. Under certain conditions, path complexes generalize simplicial complexes \cite{grigoryan_homologies_2013, grigoryan_path_2020}, thus offering a more flexible structure to work with. Even though path complexes cannot generalize regular cell complexes, we prove that our proposed Path Weisfeiler-Lehman (PWL) test is at least as powerful as CWL(\(k\)-IC) \cite{bodnar_weisfeiler_2022}, in which \(k\)-IC is a ring-based lifting map attaching 2-cells to rings of a maximum size \(k\). The realization of the PWL test via neural message-passing procedure \cite{gilmer_neural_2017} is called Path Complex Networks (PCN), which surpasses its counterparts MPSN \cite{bodnar_weisfeiler_2021} and CWN \cite{bodnar_weisfeiler_2022} in performance across various benchmarks.

\begin{figure}[t]
    \centering
    \includegraphics[width=0.7\columnwidth]{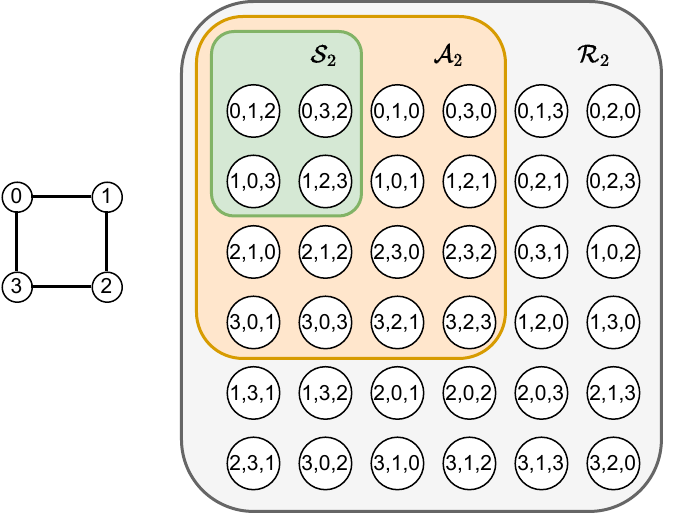}
    \caption{An illustration of different \(2\)-path spaces of a path complex arising from the graph on the left. Each element in each space is an elementary \(2\)-path that spans the corresponding space.}
    \label{fig:p_path_spaces}
\end{figure}

\paragraph{Main Contributions} Our work introduces a novel graph isomorphism test PWL and topological message-passing scheme PCN operating on path complexes, which encapsulate several theoretical properties of SWL \cite{bodnar_weisfeiler_2021} and CWL \cite{bodnar_weisfeiler_2022}. We provide theoretical connections between PWL and the latter higher-order WL tests, and we prove that PWL can indeed generalize SWL \cite{bodnar_weisfeiler_2021} and CWL(\(k\)-IC) \cite{bodnar_weisfeiler_2022}. Empirical validation of our assertions is offered through evaluations of PCN on various real-world benchmarks and a collection of 9 strongly regular graph (SRGs) families. Notably, our proposed approach achieves superior performance without the need for assumptions about graph sub-structures other than paths, which are inherent in every connected graph.

\section{Preliminaries} \label{section:prelim}

\subsection{Path Complex} \label{subsec:path-complex}

\begin{definition}{\cite{grigoryan_homologies_2013, grigoryan_path_2020}}
Given a finite non-empty set \(V\) whose element is called vertex, an \textbf{elementary p-path} on set \(V\) is any sequence of vertices with length \(p+1\). Elementary \(p\)-path is denoted by \(e_{i_0 ... i_p}\).
\end{definition}

An elementary path can be understood as the most fundamental element of a path complex. We construct a linear space defined over a field \(\mathbb{K}\) called \(\Lambda_p\) that contains all possible linear combinations of elementary \(p\)-paths. Each member of \(\Lambda_p\) is identified as a \textbf{p-path}.

\citeauthor{grigoryan_homologies_2013} also defines boundary operator  \(\partial: \Lambda_p \rightarrow \Lambda_{p-1}\) for elementary \(p\)-paths, an operator analogous to the corresponding operator for simplicial complexes.

\begin{definition}{\cite{grigoryan_homologies_2013, grigoryan_path_2020}}
Boundary operator on elementary \(p-\)paths is defined as:
\[
    \partial \, e_{i_0 ... i_p} = \sum_{q=0}^p (-1)^q e_{i_0 ... \hat{i}_q ... i_p},
\]
where \(\hat{i}_q\) indicates the removal of the index \(i_q\) from the sequence \(i_0 ... i_p\).
\end{definition}

Elementary paths may contain vertices consecutively repeated in a sequence, and these are termed as \textbf{non-regular} elementary paths. If this is not the case, the elementary paths are classified as \textbf{regular}. If we define \(\mathcal{R}_p \subset \Lambda_p\) the space encompassing all possible linear combinations of regular elementary \(p\)-paths and confine the boundary operator to \(\mathcal{R}_p\) only, the \textbf{regular boundary operator} \(\partial: \mathcal{R}_p \rightarrow \mathcal{R}_{p-1}\) will exclude any non-regular elementary \((p-1)\)-paths from the boundary set of an elementary \(p\)-path.

 \begin{definition}{\cite{grigoryan_homologies_2013, grigoryan_path_2020}}
Given a finite non-empty set \(V\), a \textbf{path complex} \(P\) is a non-empty collection of elementary paths such that for any sequence of vertices that belong to \(P\), the truncated sequences, in which either the first vertex or the last vertex is removed, are also included in \(P\).
 \end{definition}

We denote \(P_p \subset P\) where \(P_p\) contains all elementary paths with length \(p\). Elements of \(P_p\) are called \textbf{allowed elementary p-paths}, while any sequences that do not exist in \(P_p\) are called \textbf{non-allowed elementary p-paths}. Similarly, we can construct \(\mathcal{A}_p \subset \Lambda_p\) such that \(\mathcal{A}_p\) contains all possible linear combinations of allowed elementary \(p\)-paths. However, it may happen that \(\partial \, \mathcal{A}_p \not\subset \mathcal{A}_{p-1}\), as an index-omitted sequence may be non-allowed. Therefore, \citeauthor{grigoryan_homologies_2013} constructs another subspace \(\Omega_p \subseteq \mathcal{A}_p\), which is termed as \textbf{regular space of  boundary-invariant p-paths}:
\[
    \Omega_p = \{v \in \mathcal{A}_p: \partial \, v \in \mathcal{A}_{p-1}\}
\]
in which boundary operation is well-defined.

For instance, consider Figure \ref{fig:p_path_spaces} which visualizes 2-spaces associated with a path complex extended from the graph shown on the left. In this example, \(\mathcal{A}_2\) is a subset of \(\mathcal{R}_2\) due to the absence of self-loops in the graph. As illustrated in Figure \ref{fig:p_path_spaces}, none of the simple paths (no repeating vertices in a sequence) can constitute a basis for \(\Omega_2\) as there always exists a non-allowed \(1\)-path (a diagonal edge of the square) after applying boundary operation on that simple path. However, consider \(2\)-path \(v = e_{012} - e_{032}\). Applying boundary operation on \(v\):
\begin{align*}
    \partial \, v &= e_{12} - e_{02} + e_{01} - e_{32} + e_{02} - e_{03} \\
    &= e_{12} + e_{01} - e_{32} - e_{03}
\end{align*}
The outcome is a linear combination of allowed elementary \(1\)-paths, hence \(\partial \, v \in \mathcal{A}_1\). As such, \(v\) can serve as a base for \(\Omega_2\).

\subsection{Complex Lifting Transformations} \label{subsec:complex-lifting}
Given a simple graph \(G = (\mathcal{V}, \mathcal{E})\) with a finite vertex set \(\mathcal{V}\) and edge set \(\mathcal{E}\), we can apply lifting transformation such as clique complex lifting \cite{bodnar_weisfeiler_2021} or cell complex lifting \cite{bodnar_weisfeiler_2022}. Complex lifting transformations are graph pre-processing techniques such that we extend \(G\) to a complex \(K\), wherein members of \(K\) are related by part-whole relations \cite{papillon_architectures_2023} (members under a topological hierarchy) or adjacent relations (members with shared boundaries or co-boundaries). Members can be either \(k\)-simplices, \(k\)-cells, or elementary \(k\)-paths.

\begin{definition}[Boundary incidence relation]{\cite{bodnar_weisfeiler_2021, bodnar_weisfeiler_2022}}
    For any \(\sigma\) and \(\tau\) members of \(K\), \(\sigma\) is considered a boundary of \(\tau\) (denoted as \(\sigma \prec \tau\)) if and only if \(\sigma \subset \tau\) and there does not exist \(\delta \in K\) such that \(\sigma \prec \delta \prec \tau\).
\end{definition}

\begin{definition}[Relations between members]{\cite{bodnar_weisfeiler_2021, bodnar_weisfeiler_2022}} \label{def:adj_relations}
    For any member \(\sigma\) of \(K\), there are four types of relations:
    \begin{itemize}
        \item Boundary \(\mathcal{B(\sigma)} = \{\tau \ | \ \tau \prec \sigma\}\) 
        \item Co-boundary \(\mathcal{C}(\sigma) = \{\tau \ | \ \sigma \prec \tau\}\)
        \item Upper-adjacent neighborhood \(\mathcal{N}_{\uparrow}(\sigma) = \{\tau \ | \ \sigma \prec \delta \ \land \ \tau \prec \delta \}\)
        \item Lower-adjacent neighborhood \(\mathcal{N}_{\downarrow}(\sigma) = \{\tau \ | \ \delta \prec \sigma \ \land \ \delta \prec \tau \}\)
    \end{itemize}
\end{definition}

\subsection{Weisfeiler-Lehman Tests}
Weisfeiler-Lehman test (1-WL test) is a basic algorithm to investigate the isomorphism of two graphs \cite{weisfeiler_reduction_1968}. The algorithm serves as a standard method for GNNs' expressivity evaluation \cite{maron_provably_2019,xu_how_2019, morris_weisfeiler_2019,chen_can_2020,bodnar_weisfeiler_2021, bodnar_weisfeiler_2022}. 1-WL test initiates every node with the same color from a color palette, and then iteratively updates the color of each node based on its current color and the colors of its neighboring nodes. Specifically, the new color \(c_v^{(t+1)}\) of node \(v\) at time step \((t+1)\) is updated by an injective \textsc{Hash} function that maps the current color \(c_v^{(t)}\) and a collection (multi-set) of neighbors' color:
\[
    c_v^{(t+1)} = \textsc{Hash}\left(c_v^{(t)}, \left\{\!\!\left\{ c_w^{(t)}  \ | \ w \in \mathcal{N}(v) \right\}\!\!\right\}\right)
\]
The algorithm concludes when colors reach a stable state. The final graph representation is the histogram of the stable colors. If two graphs do not have the same histogram, they are not isomorphic. However, the converse does not necessarily hold true.

The algorithm proposed in \cite{grohe_pebble_2012, grohe_descriptive_2017}, which is called \(k\)-dimension Weisfeiler-Lehman test (\(k\)-WL test), is an extension of the 1-WL test that assigns colors to \(k\)-tuples of nodes \cite{morris_weisfeiler_2019, huang_short_2021}. Another variant of WL tests is \(k\)-folklore-WL test \cite{cai_optimal_1989}, which is proven to be as powerful as \((k+1)\)-WL test \cite{grohe_pebble_2012, grohe_descriptive_2017} for \(k \geq 2\).

\section{Path Weisfeiler-Lehman Test} \label{section:pwl}

\begin{figure}
    \centering
    \includegraphics[width=0.98\columnwidth]{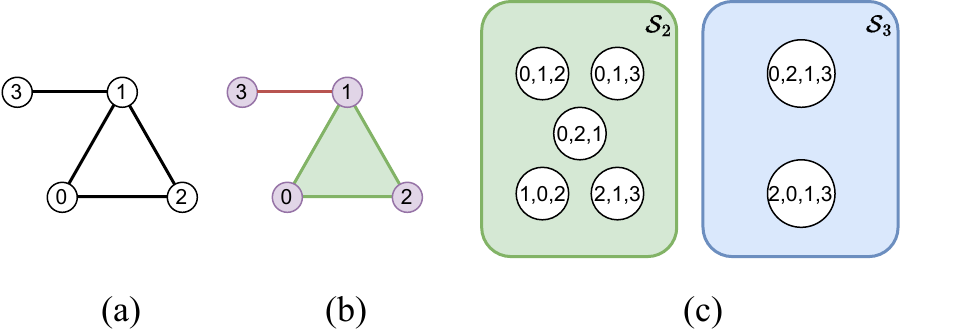}
    \caption{(a) Original graph; (b) Simplicial complex, which contains a 2-simplex, 4 1-simplices, and 4 0-simplices, arising from the original graph. Regular cell complex coincides with the simplicial complex in this case; (c) Simple path spaces \(\mathcal{S}_2\) and \(\mathcal{S}_3\) corresponding to the path complex arising from the original graph. Elementary paths of \(\mathcal{S}_0\) and \(\mathcal{S}_1\) are indeed 0-simplices (0-cells) and 1-simplices (1-cells) of the simplicial complex (regular cell complex). }
    \label{fig:complexes_comparison}
\end{figure}

\subsection{Procedure} \label{subsec:procedure}
Inspired by \cite{bodnar_weisfeiler_2021, bodnar_weisfeiler_2022}, we would like to introduce a new Weisfeiler-Lehman test based on path complex lifting transformations. However, unlike SWL\cite{bodnar_weisfeiler_2021} and CWL \cite{bodnar_weisfeiler_2022} which depend on clique or ring substructures, our method eases that prior assumption by performing color refinement solely on elementary paths. We distinguish our work with the \(k\)-set WL test proposed by \cite{morris_weisfeiler_2019}, since \(k\)-sets are non-local and may not preserve the graph's topology as elementary paths do.

\begin{figure}[t]
    \centering
    \includegraphics[width=0.9\columnwidth]{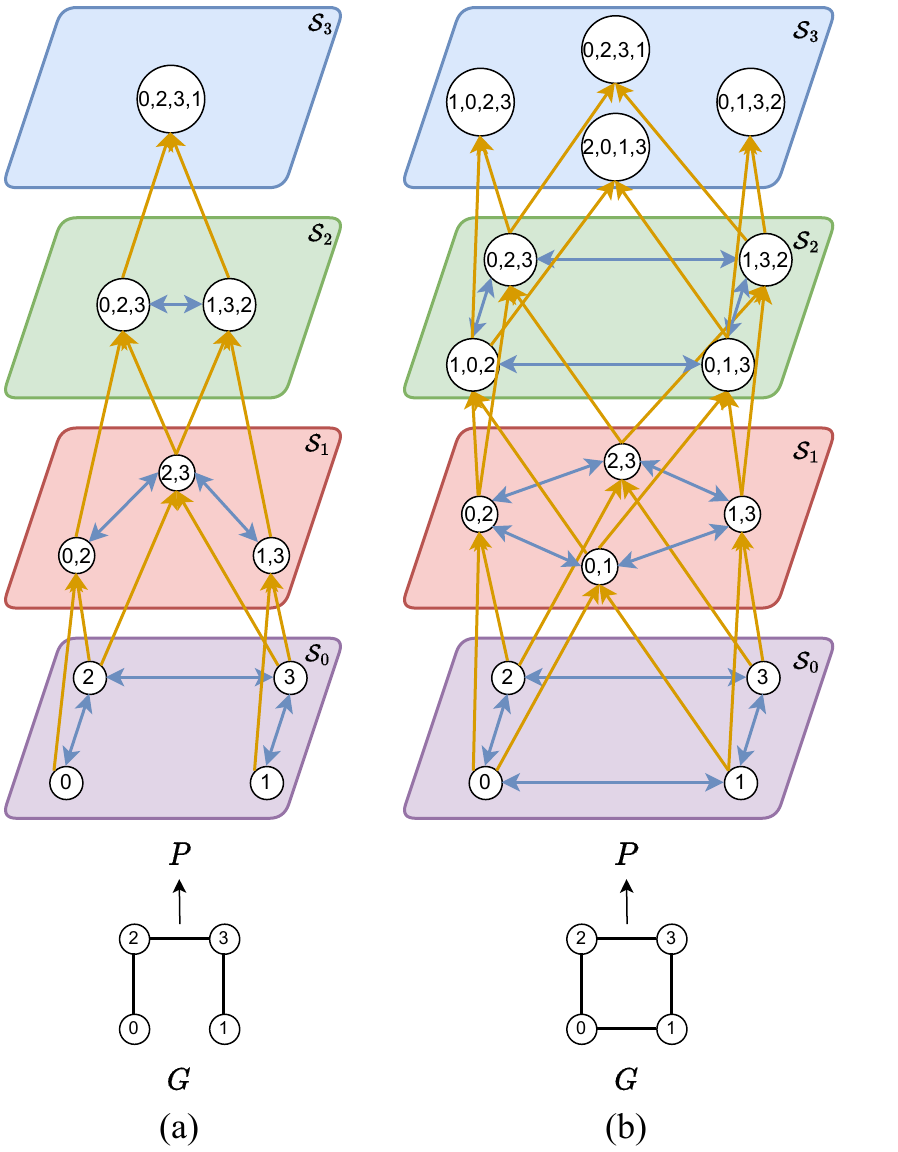}
    \caption{Examples of path complexes arising from (a) a simple path with length of 3 and (b) a ring with size of 4. Blue arrows demonstrate upper-adjacent relations, while orange arrows demonstrate boundary relations.}
    \label{fig:path_complex_example}
\end{figure}

Directly working with the regular space of boundary-invariant paths \(\Omega_p\) is non-trivial, because, as outlined in Section \ref{subsec:path-complex}, some bases that span \(\Omega_p\) are indeed formal linear combinations of elementary \(p\)-paths. Thus, we construct a new space \(\mathcal{S}_p \) where \(\mathcal{S}_p\) is spanned by all simple paths with length \(p\), which are allowed elementary \(p\)-paths of a path complex \(K\) extended by graph \(G\). Clearly,  \(\mathcal{S}_p \subset \mathcal{A}_p \, \cap \, \mathcal{R}_p\). As we only work with simple graphs, we regard any sequence and its reverse as identical. Assume there exists an injective function \(f\) that assigns each vertex in a vertex set \(V\) of a graph \(G\) a real value. Thus, we restrict \(S_p\) to only sequences where the first vertex is smaller than the last vertex. Figure \ref{fig:complexes_comparison} illustrates the main difference between our lifting transformation and the ones proposed by \cite{bodnar_weisfeiler_2021, bodnar_weisfeiler_2022}. We can observe the advantage of path complex lifting transformation, as we can theoretically lift to a higher dimension than that of simplicial complex and regular cell complex. Define \(P\) a path complex with the highest dimension \(p\) such that for any dimension \(k \leq p\), \(P_k\) contains all elementary \(k\)-paths that span \(\mathcal{S}_k\), and boundary set of any elementary \(k\)-paths is restricted to elementary \((k-1)\)-paths in \(\mathcal{S}_{k-1}\). Figure \ref{fig:path_complex_example} illustrates our idea of extending graphs to path complexes.

Drawing motivation from SWL test \cite{bodnar_weisfeiler_2021} and CWL test \cite{bodnar_weisfeiler_2022}, we define the following multi-set of colors that represents the part-whole and adjacent relations of elementary paths as defined in Definition  \ref{def:adj_relations}.

\begin{definition}[Color representations of relations]
    Given an elementary path \(\sigma\) with the associate color \(c_\sigma^t\) at time-step \(t\), we have the following multi-set of colors:
    \begin{itemize}
    \item \(c_\mathcal{B}^t(\sigma) = \left\{\!\!\left\{ c_\tau^t \, | \, \tau \in \mathcal{B}(\sigma) \right\}\!\! \right\}\)
    \item \(c_\mathcal{C}^t(\sigma) = \left\{\!\! \left\{ c_\tau^t \, | \, \tau \in \mathcal{C}(\sigma)  \right\}\!\! \right\}\)
    \item \(  c_\uparrow^t(\sigma) = \left\{\!\!\left\{  (c_\tau^t, c_\delta^t) \, | \, \tau \in \mathcal{N}_\uparrow(\sigma) \, \land \, \delta \in \mathcal{C}(\sigma, \tau)  \right\}\!\!\right\}  \)
    \item \(  c_\downarrow^t(\sigma) = \left\{\!\!\left\{  (c_\tau^t, c_\delta^t) \, | \, \tau \in \mathcal{N}_\downarrow(\sigma) \, \land \, \delta \in \mathcal{B}(\sigma, \tau)  \right\}\!\!\right\}  \)
    \end{itemize}
    where \(\mathcal{C}(\sigma, \tau) = \mathcal{C}(\sigma) \cap \mathcal{C}(\tau)\) and \(\mathcal{B}(\sigma, \tau) = \mathcal{B}(\sigma) \cap \mathcal{B}(\tau)\).
\end{definition}

Since two elementary paths may share more than one boundary or co-boundary, we adopt the coloring scheme proposed by CWL \cite{bodnar_weisfeiler_2022} as opposed to the one in SWL \cite{bodnar_weisfeiler_2021}. Path Weisfeiler-Lehman Test (PWL) performs color refinements on elementary paths that belong to \(P\) by following the below procedure:

\begin{enumerate}
    \item Initialize all elementary paths in path complex \(P\) with the same color.
    \item For every elementary path \(\sigma\), injectively map the current color of \(\sigma\) and the associated multi-sets of colors to a new color \(c_\sigma^{t+1} = \textsc{Hash}\left(c_\sigma^t, c_\mathcal{B}^t(\sigma), c_\mathcal{C}^t(\sigma), c_\downarrow^t(\sigma), c_\uparrow^t(\sigma)\right)\).
    \item Represent colorings of \(P\) by a collection of all elementary paths. Repeat Step 2 until the colorings of \(P\) are stable. Two path complexes are non-isomorphic if they have different stable color histograms of elementary paths.
\end{enumerate}

\subsection{Theoretical Results} \label{subsec:theoretical-results}
We aim to establish theoretical connections between PWL and SWL, as well as CWL. Similar to SWL and CWL, after performing \textbf{path coloring}, we compare two path complexes via \textbf{c-similarity} as proposed by \citeauthor{bodnar_weisfeiler_2021}.

\begin{definition}
    A \textbf{path coloring} is a mapping \(c\) that maps a path complex \(P\) and one of its elementary paths \(\sigma\) to a color in a fixed color palette. The mapped color is denoted by \(c_{\sigma}^P\).
\end{definition}

\begin{definition}
    Given a path coloring \(c\), two path complexes \(X\) and \(Y\) are said to be \textbf{c-similar}, denoted by \(c^X = c^Y\), if the number of elementary paths with a certain color in \(X\) is equal to the number of elementary paths with that color in \(Y\). Otherwise, \(c^X \neq c^Y\).
\end{definition}

Our paper focuses on path coloring \(c\) satisfying the condition that if \(X\) and \(Y\) are isomorphic, they are \(c\)-similar. Similar to cells in regular cell complexes, members with the same dimension in our re-defined path complexes may have different numbers of lower-dimensional elementary paths on their boundaries. Thus, we can translate the below existing theoretical results on regular cell complexes proven in \cite{bodnar_weisfeiler_2022} to path complexes. We advise the audience to read \cite{bodnar_weisfeiler_2021, bodnar_weisfeiler_2022} for detailed proofs of the following results.

\begin{definition}[Color refinement on elementary paths]\label{def:color_refinement}
    A path coloring \(c\) refines a path coloring \(d\), denoted by \(c \sqsubseteq d\), if  for all elementary paths \(\sigma\) and \(\tau\) in path complexes \(X\) and \(Y\) respectively, \(c_\sigma^X = c_\tau^Y\) implies \(d_\sigma^X = d_\tau^Y\).
\end{definition}

\begin{corollary}
    Consider two path colorings \(c\) and \(d\) on path complexes \(X\) and \(Y\) respectively such that \(c \sqsubseteq d\). If \(d^X \neq d^Y\),  then \( c^X \neq c^Y \).
\end{corollary}

\begin{theorem} \label{theorem:color refinement}
    PWL with \(\textsc{Hash}\left (c_\sigma^t, c_\mathcal{B}^t(\sigma), c_\uparrow^t(\sigma)\right )\) is as powerful as PWL with generalized update rule \(\textsc{Hash}\left (c_\sigma^t, c_\mathcal{B}^t(\sigma), c_\mathcal{C}^t(\sigma), c_\uparrow^t(\sigma), c_\downarrow^t(\sigma)\right )\).
\end{theorem}

Unless stated otherwise, we use PWL with \(\textsc{Hash}\left (c_\sigma^t, c_\mathcal{B}^t(\sigma), c_\uparrow^t(\sigma)\right )\) for the rest of the paper. 

According to \cite{grigoryan_homologies_2013, grigoryan_path_2020}, any simplicial complex can be extended naturally to a path complex if they meet two criteria: 1) The path complex should be perfect, which means any subsequence of any path must also be present in the path complex and 2) There must exist an injective real-valued vertex labeling function such that every vertex along any path must be in a monotonically increasing order. We have the following theoretical result.

\begin{theorem} \label{theorem:pwl_swl}
    PWL is at least as powerful as SWL at distinguishing non-isomorphic graphs.
\end{theorem}

Similar to cells, elementary paths can have different boundary sizes. Thus, we can also obtain the following result for PWL, which is similar to that for CWL \cite{bodnar_weisfeiler_2022}.

\begin{proposition}\label{proposition:boundary size}
    For any two path complexes \(X\) and \(Y\),  if \(\sigma \in X\) and \(\tau \in Y\) have different boundary sizes \(|\mathcal{B(\sigma)}| \neq |\mathcal{B(\tau)}|\), their colorings are different \(c_\sigma^{X,t} \neq c_\tau^{Y,t}\) for \(t > 0\).
\end{proposition}

While an elementary paths may coincide with simplex, a single elementary path could not encode a \(k\)-cell in a regular cell complex where \(k \geq 2\). For example, as shown in Figure \ref{fig:path_complex_example}, the elementary \(3\)-path \(e_{0231}\) shows up in both path complexes, but for (a), it does not coincide with \(2\)-cell if we extend the graph to a regular cell complex. However, we can prove that PWL is at least as powerful as CWL(\(k\text{-IC}\)) with the maximum dimension of 2 when distinguishing non-isomorphic simple graphs via comparing color representations of \textbf{cyclic-shifting families}.

\begin{definition}[Cyclic-shifting Family]
    Suppose \(\sigma\) is an elementary \(n\)-path of a path complex extended from a graph \(G\). If \(\sigma\) is also a \(2\)-cell of a regular cell complex extended from the same graph in which \(|\mathcal{B}(\sigma)| = n+1\), for every \(p \leq n\), there is a set of elementary \(p\)-paths which contains all sub-sequences with length \(p\) of cyclic-shifted variants of the sequence of \(\sigma\). We denote the set of such sequences \(\mathcal{F}_p(\sigma)\), which is called the cyclic-shifting \(p\)-family of \(\sigma\).
\end{definition}

For example, consider Figure \ref{fig:path_complex_example}, suppose we have a \(2\)-cell \(\sigma = e_{1023}\). The cyclic-shifting families of \(\sigma\) are:
\begin{gather*}
    \mathcal{F}_3(\sigma) = \{e_{1023}, e_{0231}, e_{0132}, e_{2013}\} \\
    \mathcal{F}_2(\sigma) = \{e_{013}, e_{023}, e_{102}, e_{132} \} \\
    \mathcal{F}_1(\sigma) = \{e_{01}, e_{02}, e_{13}, e_{23} \} \\
    \mathcal{F}_0(\sigma) = \{e_{0}, e_{1}, e_{2}, e_{3} \}
\end{gather*}
Even though \(e_{0132}\) and \(e_{2013}\) are not cyclic-shifted from \(e_{1023}\) and \(e_{0231}\), we include them instead of \(e_{2310}\) and \(e_{3102}\) in \(\mathcal{F}_3(\sigma)\) because we regard sequences and their reversed sequences identical as proposed in Section \ref{subsec:procedure}. Similarly, we obtain \(\mathcal{F}_2(\sigma), \mathcal{F}_1(\sigma)\) and \(\mathcal{F}_0(\sigma)\) as shown above.  Even though we do not explicitly take cyclic-shifting families into account when performing PWL, we can prove that cyclic-shifting families play an equivalent role to 2-cells in CWL(\(k\text{-IC}\)) via the color representations of cyclic-shifting families.

\begin{definition}[Color Representation of Cyclic-shifting Family]
    For any path complex \(X\), given an elementary \(p\)-path \(\sigma \in X\) and \(\sigma\) also coincides with 2-cells with the boundary size of \(p+1\), the color representation of \(\mathcal{F}_p(\sigma)\) at time-step \(t\) is denoted as a multi-set:
\[
    C_p^{X, t}(\sigma) = \left\{\!\!\left\{ c^{X, t}_{\delta_\sigma} \, | \, \delta_\sigma \in \mathcal{F}_p(\sigma) \right\}\!\!\right\} 
\]
\end{definition}
Then, we have the following proposition.

\begin{proposition}\label{proposition:cyclic-shifting families}
    For any two path complexes \(X\) and \(Y\), if two cyclic-shifting \(p\)-families have similar color representation  \( C_p^{X, t + p}(\sigma) = C_p^{Y, t + p}(\tau)\) in which \(\sigma \in X\) and \(\tau \in Y\), cyclic-shifting \(k\)-families also have similar color representation \(C_k^{X, t + k}(\sigma) = C_k^{Y, t + k}(\tau)\) where \(k \leq p\).
\end{proposition}

Proposition \ref{proposition:cyclic-shifting families}  implies that similar higher-order cyclic-shifting families can induce similar lower-order cyclic-shifting families. This is particularly relevant for 1-families, which include 1-paths and 1-cells. We have the following lemmas.

\begin{lemma} \label{lemma:p-paths to 0-cells}
    For any pair of path complexes \(X\) and \(Y\) arising from two graphs \(G_1\) and \(G_2\), if two elementary 0-paths (1-paths whose co-boundaries are not belong to any cyclic-shifting 2-families) \(\sigma \in X\) and \(\tau \in Y\) have similar colorings \(c_\sigma^{X,t} = c_\tau^{Y,t}\) performed by PWL at time-step \(t\), we have two corresponding 0-cells (1-cells not adjacent to any 2-cells) \(\sigma \in A\) and \(\tau \in B\) with similar color representations \(d_\sigma^{A,t} = d_\tau^{B,t}\) performed by CWL(\(k\)-IC) at time-step \(t\), where \(A\) and \(B\) are regular cell complexes arising from \(G_1\) and \(G_2\).
\end{lemma}

\begin{lemma}\label{lemma:p-paths to 2-cells}
    For any two path complexes \(X\) and \(Y\) arising from two graphs \(G_1\) and \(G_2\), if cyclic-shifting \((n-1)\)-families of \(\sigma \in X\) and \(\tau \in Y\)  have similar color representations \(C_{n-1}^{X,t + n - 3}(\sigma) = C_{n-1}^{Y, t + n - 3}(\tau)\) performed by PWL at time-step \(t + n - 3\), we have two corresponding 2-cells \(\alpha \in A\) and \(\beta \in B\) with similar colorings \(d_\alpha^{A,t} = d_\beta^{B,t}\) performed by CWL(\(k\)-IC) at time-step \(t\), in which \(A\) and \(B\) are regular cell complexes arising from \(G_1\) and \(G_2\) and \(n\) is the boundary size of the 2-cells.
\end{lemma}

\begin{lemma} \label{lemma:p-paths to 1-cells}
    For any two path complexes \(X\) and \(Y\) arising from two graphs \(G_1\) and \(G_2\), if two elementary 1-paths \(\alpha \in X\) and \(\beta \in Y\) have the same coloring \(c_\alpha^{X,t} = c_\beta^{Y,t}\) and their cyclic-shifting \((n-1)\)-families have similar color representations \(C_{n-1}^{X,t + n - 3}(\sigma) = C_{n-1}^{Y, t + n - 3}(\tau)\) performed by PWL at time-step \(t + n - 3\), we have two corresponding 1-cells \(\alpha \in A\) and \(\beta \in B\) with the same coloring \(d_\alpha^{A,t} = d_\beta^{B,t}\) performed by CWL(\(k\)-IC) at time-step \(t\), in which \(A\) and \(B\) are regular cell complexes arising from \(G_1\) and \(G_2\) and \(n\) is the boundary size of the 2-cells that are adjacent to \(\alpha\) and \(\beta\).
\end{lemma}

Given the above lemmas, as all cells with different orders of 2-dimensional regular cell complexes can be induced from path complexes and their corresponding cyclic-shifting families, we come up with the following theorem and corollaries.

\begin{theorem} \label{theorem:pwl_cwl}
    PWL is at least as powerful as CWL(\(k\text{-IC}\)) with the maximum dimension of 2 at distinguishing non-isomorphic graphs.
\end{theorem}

\begin{corollary}
    \label{corollary:pwl_wl}
    PWL is strictly more powerful than WL at distinguishing non-isomorphic graphs.
\end{corollary}

\begin{corollary}
    \label{corollary:pwl_3wl}
    PWL is not less powerful than 3-WL at distinguishing non-isomorphic graphs.
\end{corollary}

\section{Realization of PWL Test via Message-passing Neural Networks} \label{section:pcn}

Similar to \cite{bodnar_weisfeiler_2021, bodnar_weisfeiler_2022}, a natural way to realize PWL is through Message Passing Neural Networks (MPNNs) \cite{gilmer_neural_2017}, in which MPNNs pass messages along elementary paths based on part-whole and adjacent relations as defined in Section \ref{subsec:complex-lifting}. As demonstrated by Theorem \ref{def:color_refinement} proven by \citeauthor{bodnar_weisfeiler_2022}, we can exclude the message from co-boundaries and lower-adjacent neighbors. Thus, we update feature of an elementary path by the following:
\begin{gather*}
    m_{\mathcal{B}}^{(t+1)}(\sigma) = \text{AGG}_{\tau \in \mathcal{B}(\sigma)} \left( \text{M}_{\mathcal{B}} (h_{\sigma}^{(t)}, h_{\tau}^{(t)})  \right) \\
     m_{\uparrow}^{(t+1)}(\sigma) = \text{AGG}_{\tau \in \mathcal{N}_{\uparrow}(\sigma), \delta \in \mathcal{C}(\sigma, \tau)} \left( \text{M}_{\uparrow} (h_{\sigma}^{(t)}, h_{\tau}^{(t)}, h_{\delta}^{(t)})  \right) \\
     h_{\sigma}^{(t+1)} = \text{UP} \left(h_{\sigma}^{(t)}, m_{\mathcal{B}}^{(t)}(\sigma), m_{\uparrow}^{(t+1)}(\sigma) \right) 
\end{gather*}
in which \(\text{M}_{\mathcal{B}}, \text{M}_{\mathcal{\uparrow}}\) and \(\text{UP}\) can be modeled by learnable functions. PCN inherit all properties of MPSN \cite{bodnar_weisfeiler_2021} and CWN \cite{bodnar_weisfeiler_2022}, given that they all operate under the same mechanism. Thus, we naturally have the following results, whose proofs are identical to those in \cite{bodnar_weisfeiler_2022}.

\begin{theorem}
    PCNs are at most powerful as PWL. PCNs can be as powerful as PWL if PCNs are equipped with a sufficient number of layers and injective aggregators.
\end{theorem}

\begin{theorem}
    PCN layers are elementary path permutation equivariant.
\end{theorem} 
For the rest of the paper, we leverage the following model called Path Isomorphism Network (PIN). The architecture of the model is identical to SIN \cite{bodnar_weisfeiler_2021} and CIN \cite{bodnar_weisfeiler_2022}, albeit with slight alterations. The main distinguishing factor lies in the graph lifting transformation. This family of models is commonly regarded as a higher-order version of GIN \cite{xu_how_2019} because of the similar update formula. The update formulae are detailed in Appendix \ref{app:formula-mp}.

\section{Experiments} \label{section:experiments}

We evaluate PIN on several real-world datasets across different domains such as molecular graphs or social graphs. We also provide an in-depth empirical study, which supports our claim that PWL is at least as powerful as CWL(\(k\)-IC), on SRGs. Detailed hyperparameter settings along with relevant ablation studies are documented in Appendix \ref{app:experiments}.

\begin{table}[t]
\centering
\resizebox{\columnwidth}{!}{%
\begin{tabular}{@{}lcccc@{}}
\toprule
Dataset     & PROTEINS   & NCI1       & NCI109     & IMDB-B     \\ \midrule
RWK        & 59.6 ± 0.1 & \(>\) 3 days    & N/A        & N/A        \\
GK (k=3)   & 71.4 ± 0.3 & 62.5 ± 0.3 & 62.4 ± 0.3 & N/A        \\
PK         & 73.7 ± 0.7 & 82.5 ± 0.5 & N/A        & N/A        \\
WL Kernel  & 75.0 ± 3.1 & 86.0 ± 1.8 \(\blacklozenge\) & N/A        & 73.8 ± 3.9 \\ \midrule
DCNN     & 61.3 ± 1.6 & 56.6 ± 1.0 & N/A        & 49.1 ± 1.4 \\
DGCNN     & 75.5 ± 0.9 & 74.4 ± 0.5 & N/A        & 70.0 ± 0.9 \\
IGN      & 76.6 ± 5.5 & 74.3 ± 2.7 & 72.8 ± 1.5 & 72.0 ± 5.5 \\
GIN     & 76.2 ± 2.8 & 82.7 ± 1.7 & N/A        & 75.1 ± 5.1 \\
PPGNs      & 77.2 ± 4.7 & 83.2 ± 1.1 & 82.2 ± 1.4 & 73.0 ± 5.8 \\
Natural GN  & 71.7 ± 1.0 & 82.4 ± 1.3 & N/A        & 73.5 ± 2.0 \\
GSN      & 76.6 ± 5.0 & 83.5 ± 2.0 & N/A        & 77.8 ± 3.3 \(\blacklozenge\) \\
pathGCN   & 80.4 ± 4.2 \(\blacktriangle\) & 83.3 ± 1.3 & N/A        & N/A        \\
 PathNN& 75.2 ± 3.9& 82.3 ± 1.9& N/A&72.6 ± 3.3\\ \midrule
SIN \textsuperscript{\textdagger}       & 76.4 ± 3.3 & 82.7 ± 2.1 & N/A        & 75.6 ± 3.2 \(\bullet\) \\
CIN  \textsuperscript{\textdagger}       & 77.0 ± 4.3 & 83.6 ± 1.4 & 84.0 ± 1.6 \(\bullet\) & 75.6 ± 3.7 \\
CAN       & 78.2 ± 2.0 & 84.5 ± 1.6 & 83.6 ± 1.2 & N/A        \\
CIN++  & 80.5 ± 3.9 \(\blacklozenge\) & 85.3 ± 1.2 \(\blacktriangle\) & 84.5 ± 2.4 \(\blacklozenge\) & N/A        \\ \midrule
PIN (Ours)        & 78.8 ± 4.4 \(\bullet\) & 85.1 ± 1.5 \(\bullet\) & 84.0 ± 1.5 \(\blacktriangle\) & 76.6 ± 2.9 \(\blacktriangle\) \\ \bottomrule
\end{tabular}%
}
\caption{TUDataset Benchmarks. The first part consists of graph kernel methods, the second part consists of GNNs, and the third part consists of higher-order GNNs. The top-3 methods in each benchmark are denoted by \(\blacklozenge\) (\(1^{\text{st}}\) place),  \(\blacktriangle\) (\(2^{\text{nd}}\) place), and \(\bullet\) (\(3^{\text{rd}}\) place). Baselines are denoted by \textdagger.}
\label{tab:tu-datasets}
\end{table}

\begin{table*}[t]
\centering
\fontsize{10}{11}\selectfont
\resizebox{0.574\textwidth}{!}{%
\begin{tabular}{@{}lcccc@{}}
\toprule
\multirow{2}{*}{Dataset} & \multicolumn{2}{c}{ZINC}                        & \multicolumn{2}{c}{OGBG-MOLHIV}                    \\ \cmidrule(l){2-5} 
                         & No Edge Feat.                  & W/  Edge Feat.                  & Test ROC-AUC          & Val. ROC-AUC          \\ \midrule
GCN                      & 0.469 ± 0.002          & N/A                    & N/A                   & N/A                   \\
GAT                     & 0.463 ± 0.002          & N/A                    & N/A                   & N/A                   \\
GatedGCN               & 0.422 ± 0.006          & 0.363 ± 0.009          & N/A                   & N/A                   \\
GIN                    & 0.408 ± 0.008          & 0.252 ± 0.014          & 77.07 ± 1.49          & 84.79 ± 0.68          \\
PNA                 & 0.320 ± 0.032          & 0.188 ± 0.004          & 79.05 ± 1.32          & 85.19 ± 0.99          \\
DGN                     & 0.219 ± 0.010          & 0.168 ± 0.003          & 79.70 ± 0.97          & 84.70 ± 0.47          \\
HIMP                & N/A                    & 0.151 ± 0.006          & 78.80 ± 0.82          & N/A                   \\
GSN                  & 0.140 ± 0.006          & 0.115 ± 0.012          & 77.99 ± 1.00          & \textbf{86.58 ± 0.84} \\
 PathNN& N/A& 0.090 ± 0.004& 79.17 ± 1.09&N/A\\
CIN \textsuperscript{\textdagger}                     & \textbf{0.115 ± 0.003} & 0.079 ± 0.006          & \textbf{80.94 ± 0.57} & N/A                   \\
CIN++               & N/A                    & \textbf{0.077 ± 0.004} & 80.63 ± 0.94          & N/A                   \\ \midrule
PIN (Ours)               & 0.139 ± 0.004          & 0.096 ± 0.006          & 79.44 ± 1.40          & 82.41 ± 0.96           \\
\bottomrule
\end{tabular}%
}
\caption{ZINC and OGBG-MOLHIV datasets. Bold texts indicate the best performance. Performance on ZINC is evaluated by Mean Squared Error, while performance on OGBG-MOLHIV is evaluated by ROC-AUC. Baseline is denoted by \textdagger.}
\label{tab:zinc-molhiv}
\end{table*}

\subsection{TUDataset Benchmarks} \label{subsec:tudataset}
TUDataset benchmarks, encompassing a broad spectrum of graph datasets from biology, chemistry, and social networks, are proposed in \cite{morris_tudataset_2020}. In this paper, we evaluate our model on 4 different benchmarks on classification tasks: PROTEINS, NCI1, NCI109, and IMDB-B. However, an evaluation of our model on large graphs such as those in RDT-B is unattainable due to the prohibitive time complexity. The evaluation methodology adheres to the procedure outlined by  \cite{xu_how_2019}. Specifically, we report the highest mean test accuracy across a 10-fold cross-validation as indicated in \cite{xu_how_2019}.

As illustrated in Table \ref{tab:tu-datasets}, an updated version of the tables presented in \cite{bodnar_weisfeiler_2022, giusti_cin_2023}, PIN exhibits superior accuracy in comparison to SIN \cite{bodnar_weisfeiler_2021} and CIN \cite{bodnar_weisfeiler_2022}, even on molecular graph datasets, where rings hold considerable significance.

For the TUDataset Benchmarks, we compare performance of three types of methods: graph kernels (RWK \cite{gartner_on_graph_2003}, GK \cite{shervashidze_efficient_2009}, PK \cite{neumann_propagation_2016}, WL Kernel \cite{shervashidze_wl_kernels_2011}), GNNs (DCNN \cite{atwood_diffusion_convolution_2016}, DGCNN \cite{zhang_end_to_end_2018}, IGN \cite{maron_invariant_2019}, GIN \cite{xu_how_2019}, PPGNs \cite{maron_provably_2019}, Natural GN \cite{dehaan_natural_2020}, GSN \cite{bouritsas_improving_2021}, pathGCN \cite{eliasof_pathgcn_2022}, PathNN \cite{michel_expressive_2023}) and higher-order GNNs (SIN \cite{bodnar_weisfeiler_2021}, CIN \cite{bodnar_weisfeiler_2022}, CAN \cite{giusti_cell_2022}, CIN++ \cite{giusti_cin_2023}).

\begin{figure*}[t]
    \centering
    \includegraphics[width=\textwidth]{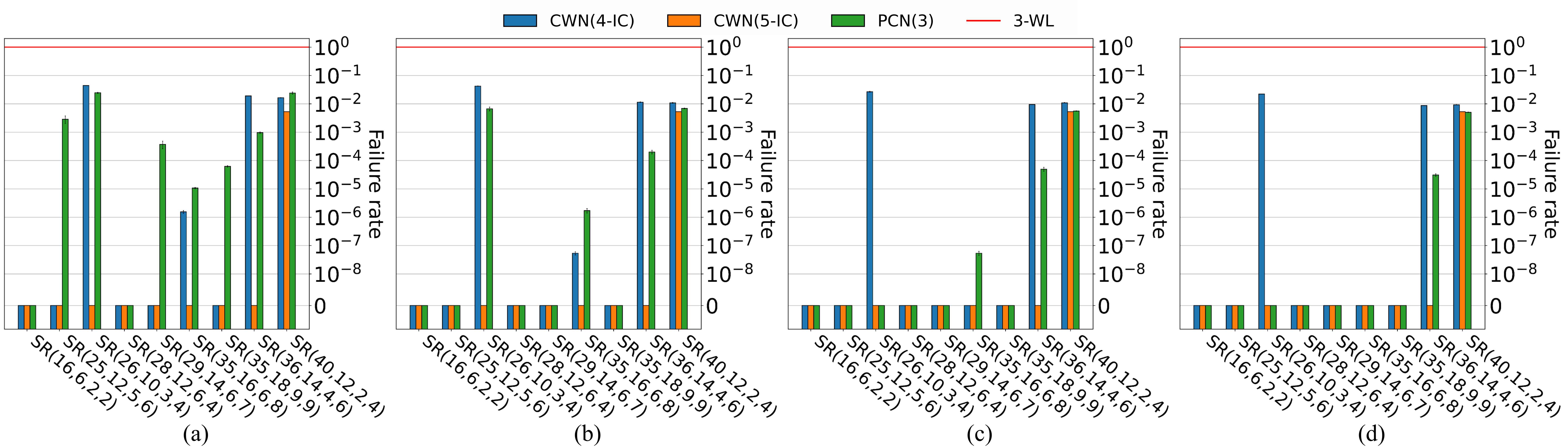}
    \caption{Failure rate comparison between CWN and PCN on SRG Families over 10 different seeds. (a) 3 message-passing layers. (b) 4 message-passing layers. (c) 5 message-passing layers. (d) 6 message-passing layers.}
    \label{fig:sr-graphs}
\end{figure*}

\subsection{ZINC} \label{subsec:zinc}

We also evaluate our model on the ZINC dataset, a common graph benchmark for regression tasks \cite{sterling_zinc15_2015,dwivedi_benchmarkgnns_2020}. As shown in Table \ref{tab:zinc-molhiv}, which is an updated table of \cite{bouritsas_improving_2021} and \cite{giusti_cin_2023}, our model's performance is lower than that of the baseline CIN \cite{bodnar_weisfeiler_2022} on the test set of ZINC. Our conjecture regarding this performance gap is primarily attributed to the low dimensionality of the path complex, which cannot fully represent large ring structures, and the shallowness of our architectures, which cannot propagate messages efficiently from elementary 0-paths to the highest-dimensional elementary paths. However, our model outperforms the remaining state-of-the-art methods by large margins.

For the ZINC dataset, we compare performance of GCN \cite{kipf_semi-supervised_2017}, GAT \cite{velickovic_graph_2018}, GatedGCN \cite{bresson_residual_2018}, GIN \cite{xu_how_2019}, PNA \cite{corso_pna_2020}, DGN \cite{beaini_directional_2020}, HIMP \cite{fey_hierarchical_2020}, GSN \cite{bouritsas_improving_2021}, PathNN \cite{michel_expressive_2023}, CIN \cite{bodnar_weisfeiler_2022}, and CIN++ \cite{giusti_cin_2023}.

\subsection{OGBG-MOLHIV} \label{subsec:molhiv}

We also evaluate our model on the OGBG-MOLHIV dataset \cite{hu_ogb_2020}, which contains about 41k graphs for the graph binary classification task. \citeauthor{hu_ogb_2020} applies a scaffold splitting procedure \cite{wu_moleculenet_2018} for the OGBG-MOLHIV dataset; thus we report the performance of our model on both the validation and test sets in Table \ref{tab:zinc-molhiv}. Analogous to the results observed from the ZINC dataset, our model performs less effectively than the baselines. For OGBG-MOLHIV, we compare our model to the same methods as in ZINC.

\subsection{Strongly Regular Graphs} \label{subsec:sr-graphs}
The original SRG data is publicly available\footnote{http://users.cecs.anu.edu.au/\raisebox{0.5ex}{\texttildelow}bdm/data/graphs.html}. Similar to \cite{bodnar_weisfeiler_2021, bodnar_weisfeiler_2022, bouritsas_improving_2021}, we adopt the SRG benchmark, which poses as hard examples for the task of distinguishing non-isomorphic graphs, to support our theory behind PWL and PCN. Figure \ref{fig:sr-graphs} illustrates the performance of PCN when lifting graphs to a dimension of 3, which is denoted by PCN(3). As we increase the number of layers for PCN, PCN outperforms CWN(4-IC), and achieves an on-par performance with CWN(5-IC), even though we do not explicitly include the inductive bias of ring-shaped structures. This result aligns with our theory that PWL can generalize CWL(\(k\)-IC) if the highest-order elementary \(p\)-paths align with the \(2\)-cells with boundary size of \(p+1\), which can be empirically validated by comparing CWN(4-IC) and PCN(3). The result also illustrates that PWL takes a longer time to reach stable colorings than CWN does, as messages between two further edges of a ring must be passed through all dimensions of a path complex.

\section{Discussion and Related Works}

\paragraph{Higher-order GNNs} Topological deep learning, which investigates beyond pair-wise interactions between vertices, has gained significant attention lately. \citeauthor{papillon_architectures_2023} and \citeauthor{hajij_topological_2023} recently conducted a survey on this field, and classify learning domains as either part-whole relations (simplicial complex \cite{bodnar_weisfeiler_2021, ebli_simplicial_2020, roddenberry_principled_2021} and cellular complex \cite{bodnar_weisfeiler_2022, giusti_cin_2023}), set-type relations (hypergraphs \cite{huang_unignn_2021, feng_hypergraph_2018, yadati_hypergcn_2019}) or both (combinatorial complex \cite{hajij_topological_2023}). Some methods such as \cite{maron_provably_2019, morris_weisfeiler_2019} are provably as powerful as the \(k\)-WL test \cite{grohe_pebble_2012, grohe_descriptive_2017}, or even strictly more expressive \cite{morris_weisfeiler_2020}. However, these approaches carry a greater computational cost than ours, due to the necessity of \(k\)-tuple inputs, leading to potential loss of locality and an increased risk of overfitting. \citeauthor{bodnar_weisfeiler_2021} introduce message-passing models over simplicial complexes and cellular complexes, which are provably not less powerful than the 3-WL test. Our learning domain falls into part-whole relations, and our proposed model is proven to generalize SWL \cite{bodnar_weisfeiler_2021} and CWL(\(k\)-IC) \cite{bodnar_weisfeiler_2022} while still retaining locality.

\paragraph{Path-based Graph Learning}  Paths constitute a vital component of all graphs, providing valuable insights into the graphs. pathGCN \cite{eliasof_pathgcn_2022} proposes to learn spatial operators from randomly sampled paths as a way to encode multi-level local neighborhoods. PathNN \cite{michel_expressive_2023} updates a node representation based on paths with different lengths passing through that node. GSN \cite{bouritsas_improving_2021} explicitly encodes nodes or edges by counting the numbers of certain sub-structures that nodes or edges belong to. \citeauthor{bouritsas_improving_2021} empirically finds that counting paths with the maximum length of 6 is enough to distinguish all graphs in the SRG families; while counting paths with the maximum length of 3 is insufficient to distinguish any graphs. Our experiment shows that our model can tell apart almost every SRG family with a dimension of 3, which empirically proves that message passing between elementary paths have greater capabilities beyond path counting.

\paragraph{Limitations} Similar to \cite{bodnar_weisfeiler_2021, bodnar_weisfeiler_2022}, our approach incurs high time and space complexities. A naive way to enumerate all paths with length \(k\) for a graph with \(n\) nodes has worst time complexity of \(\mathcal{O}(n^{k+1})\). If we take the branching factor \(b\) of the graph into account, the worst time complexity is \(\mathcal{O}(nb^k)\)\cite{michel_expressive_2023}. In practice, the majority of graph datasets that we evaluated are sparse, and lifting them to the third dimension, or even higher, is feasible. The only exceptions are PROTEINS, IMDB-B, and SR(35,16,6,8). However, with path complexes with a dimension of only 3, we can achieve superior performance than our baselines on the SRG experiments even though theoretically we cannot represent the rings with a size above 4. For PROTEINS and IMDB-B, these datasets are relatively small (about 1,000 graphs), and thus lifting to high dimensions may cause overfitting. Unlike CWL \cite{bodnar_weisfeiler_2022} where the furthest edges on the boundary of a ring can exchange messages after 2 propagations, PWL requires passing messages on every order of a path complex. Yet, increasing the number of message-passing layers can lead to over-smoothing \cite{oono_graph_2021, li_deeper_2018}. Path complex can be visualized as  \textsc{Tree-NeighborsMatch} problem proposed in \cite{alon_bottleneck_2021}, where each node can be considered as an elementary \(p\)-path having at least two children nodes, which are elementary \((p-1)\)-paths on the boundary. Thus, over-squashing may occur for high-dimensional path complexes. 

\paragraph{Future Works}  High-dimensional elementary paths are more prevalent and trivial than their counterparts, which are simplices and cells. However, high-dimensional path complexes are prone with over-smoothing \cite{oono_graph_2021, li_deeper_2018} and over-squashing \cite{alon_bottleneck_2021}, which have not yet been encountered in other topological domains. An in-depth study into these problems could unlock advancements in higher-order graph augmentation techniques such as dropout \cite{rong_dropedge_2020, papp_dropgnn_2021} or graph rewiring \cite{topping_understanding_2022}. PWL and PWN rely on spaces of simple paths \(\mathcal{S}_p\), while the original path complex is well-defined on the regular space of boundary-invariant \(\Omega_p\), which suggests there may exist a more powerful representation of path complexes.

\section{Conclusion}

We have demonstrated how path complexes can be an alternative topological domain for simplicial and regular cell complexes. We verify, theoretically and experimentally, the generalization of path complexes over the other two topological domains. The versatility of our approach, without the necessity for inductive bias on graph substructures, underscores the centrality and universality of paths as the foundational elements of any graph.

\bibliography{aaai24}

\clearpage
\appendix
\section{Proofs}

\subsection{Proof for Theorem \ref{theorem:pwl_swl}}

\begin{proof}
    The high-level idea is that every simplicial complex is a path complex if the two criteria in Section \ref{subsec:theoretical-results} are satisfied. Thus, PWL always induces SWL. The proof is however similar to that in \cite{bodnar_weisfeiler_2022}, and we advise readers to review the proof in \cite{bodnar_weisfeiler_2022}. 
\end{proof}

\subsection{Proof for Proposition \ref{proposition:boundary size}}

\begin{proof}
    It is obvious that after one iteration, \(c_\sigma^{X,1} \neq c_\tau^{Y,1}\) because \(c_\mathcal{B}^{X,0}(\sigma) \neq c_\mathcal{B}^{Y,0}(\tau)\).
\end{proof}

\subsection{Proof for Proposition \ref{proposition:cyclic-shifting families}}

\begin{proof}
     Let \(c\) be the path coloring performed by PWL. Consider color representations of the cyclic-shifting families of elementary \(p\)-paths \(\sigma \in X\) and \(\tau \in Y\) at an arbitrary time-step \(t+p\):
    \begin{gather*}
    C_{p}^{X, t + p}(\sigma) = \left\{\!\!\left\{ c^{X, t+p}_{\delta_\sigma} \, | \, \delta_\sigma \in \mathcal{F}_p(\sigma) \right\}\!\!\right\} \\
    C_{p}^{Y, t + p}(\tau) = \left\{\!\!\left\{ c^{Y, t+p}_{\delta_\tau} \, | \, \delta_\tau \in \mathcal{F}_p(\tau) \right\}\!\!\right\}
  \end{gather*}
We assume the highest dimension of the cyclic-shifting families is at least 3. The special case where the highest dimension of the cyclic-shifting families is 2, which also means the corresponding 2-cell is a triangle, can be easily treated separately. Expanding the update \textsc{Hash} function and only considering the colorings of the boundary, we have the following equality:
\[
\left\{\!\!\left\{ c^{X, t+p-1}_\mathcal{B}(\delta_\sigma)   \right\}\!\!\right\} = \left\{\!\!\left\{c^{Y, t+p-1}_\mathcal{B}(\delta_\tau)   \right\}\!\!\right\} \]
For each \(\delta_\sigma\) and \(\delta_\tau\), we have:
\begin{gather*}
 c^{X, t+p-1}_\mathcal{B}(\delta_\sigma) = \left\{\!\!\left\{ c^{X,t+p-1}_{\psi_\sigma} \, | \, \psi_\sigma \in \mathcal{B}(\delta_\sigma) \right\}\!\!\right\} \\
 c^{Y, t+p-1}_\mathcal{B}(\delta_\tau) = \left\{\!\!\left\{ c^{X,t+p-1}_{\psi_\tau} \, | \, \psi_\tau \in \mathcal{B}(\delta_\tau)  \right\}\!\!\right\} 
\end{gather*}
where \(\psi_\sigma \in \mathcal{F}_{p-1}(\sigma)\) and \(\psi_\tau \in \mathcal{F}_{p-1}(\tau)\). Thus, we have:
\[
\left\{\!\!\left\{ c^{X, t+p-1}_{\psi_\sigma} \right\}\!\!\right\} = \left\{\!\!\left\{c^{Y, t+p-1}_{\psi_\tau}\right\}\!\!\right\} \]
where each elementary \((p-1)\)-path shows up twice. Re-winding back to time-step \(t + k\), we will have:
\[
\left\{\!\!\left\{ c^{X, t + k}_{\gamma_\sigma} \right\}\!\!\right\} = \left\{\!\!\left\{c^{Y, t + k}_{\gamma_\tau}\right\}\!\!\right\} \]
where each elementary \(k\)-path \(\gamma_\sigma \in \mathcal{F}_{k}(\sigma)\) and \(\gamma_\tau \in \mathcal{F}_{k}(\tau)\) shows up with a multiplicity of \(2^{(p-k)}\). Removing other \(2^{(p-k)} - 1\) duplicates for each elementary \(k\)-path, we have:
\[
    C_k^{X, t + k}(\sigma) = C_k^{Y, t + k}(\tau)
\]
The proof for the special case is similar, except that each elementary 1-path shows up three times instead (because all three elementary 1-paths are allowed) when re-winding one time-step.
\end{proof}
\subsection{Proof for Lemma \ref{lemma:p-paths to 0-cells}}

\begin{proof}
    PWL is identical to CWL for dimension 0 because elementary 0-path and 0-cell have the same update \textsc{Hash} function. For dimension 1 where we consider only elementary 1-paths whose co-boundaries are not belong to any cyclic-shifting 2-families, and 1-cells that are not adjacent to any 2-cells, the update \textsc{Hash} function of the former contains arguments for that of the latter.
\end{proof}
\subsection{Proof for Lemma \ref{lemma:p-paths to 2-cells}}

\begin{proof}
    Let \(c\) be the path coloring performed by PWL and \(d\) be the cell coloring performed by CWL.
    Without losing generality, consider rings with an arbitrary size \(n \geq 3\). Assume that given two c-similar path complexes, there always exists a bijective mapping that maps cyclic-shifting families of the path complexes such that their color representations are similar. We would like to prove \(C_{n-1}^{X,t + n - 3}(\sigma) = C_{n-1}^{Y, t + n - 3}(\tau)\) implies \(d^{A,t}_\alpha = d^{B,t}_\beta\) where \(\alpha \in \mathcal{F}_{n-1}(\sigma)\) and \(\beta \in \mathcal{F}_{n-1}(\tau)\) are also \(2\)-cells if we extend the original graphs to regular cell complexes.
    
    We would like to prove this by induction. The base case obviously holds because, at time-step \(t = 0\), every cell is assigned the same color. Suppose the above statement holds for \(t\). We would like to prove it also holds for \(t+1\), which is equivalent to that given \(C_{n-1}^{X,t + n - 2}(\sigma) = C_{n-1}^{Y, t + n - 2}(\tau)\), prove \(d^{A,t+1}_\alpha = d^{B,t+1}_\beta\). By Proposition \ref{proposition:cyclic-shifting families},  \(C_{n-1}^{X,t + n - 2}(\sigma) = C_{n-1}^{Y, t + n - 2}(\tau)\) implies \(C_{1}^{X,t}(\sigma) = C_{1}^{Y, t}(\tau)\). Because elementary 1-path is 1-cell, we have the boundary adjacencies used by CWL at time-step \(t\):
\[
d_\mathcal{B}^{A,t}(\alpha) = d_\mathcal{B}^{B,t}(\beta)
\]
where \(\alpha\) and \(\beta\) are 2-cells in \(A\) and \(B\) and also elementary \((n-1)\)-paths in \(\mathcal{F}_{n-1}(\sigma)\) and \(\mathcal{F}_{n-1}(\tau)\).
Also, by the induction hypothesis, we also have \(d^{A,t}_\alpha = d^{B,t}_\beta\). As 2-cells do not have upper adjacencies if we leverage ring-based lifting transformation with a maximum dimension of 2, we conclude that \(d^{A,t+1}_\alpha = d^{B,t+1}_\beta\).
\end{proof}

\subsection{Proof for Lemma \ref{lemma:p-paths to 1-cells}}

\begin{proof}
    Similar to the proof for Lemma \ref{lemma:p-paths to 2-cells}, we use the same notations for different colorings. We also assume that given two c-similar path complexes, there always exists a bijective mapping that maps cyclic-shifting families of the path complexes such that their color representations are similar.
    
    Without losing generality, consider rings with an arbitrary size \(n \geq 3\). We would like to prove that \(c_\alpha^{X,t} = c_\beta^{Y,t}\) and \(C_{n-1}^{X,t + n - 3}(\sigma) = C_{n-1}^{Y, t + n - 3}(\tau)\) imply \(d_\alpha^{A,t} = d_\alpha^{B,t}\) by induction.
    
    The base case obviously holds because, at time-step \(t = 0\), every cell is assigned the same color. Suppose the above statement is true for \(t\). We would like to prove that \(c_\alpha^{X,t+1} = c_\beta^{Y,t+1}\) and \(C_{n-1}^{X,t + n - 2}(\sigma) = C_{n-1}^{Y, t + n - 2}(\tau)\) imply \(d_\alpha^{A,t+1} = d_\beta^{B,t+1}\).
    
    By induction hypothesis, we have \(d_\alpha^{A,t} = d_\alpha^{B,t}\). By Lemma \ref{lemma:p-paths to 0-cells}, \(c_\mathcal{B}^{X,t}(\alpha) = c_\mathcal{B}^{Y,t}(\beta)\), which can be obtained from \(c_\alpha^{X,t+1} = c_\beta^{Y,t+1}\), implies \(d_\mathcal{B}^{A,t}(\alpha) = d_\mathcal{B}^{B,t}(\beta)\) because elementary 0-paths (0-cells) are on the boundary of elementary 1-paths (1-cells). By Lemma \ref{lemma:p-paths to 2-cells}, \(C_{n-1}^{X,t + n - 2}(\sigma) = C_{n-1}^{Y, t + n - 2}(\tau)\) implies  \(d_\delta^{A,t+1} = d_\gamma^{B,t+1}\) in which \(\alpha \prec \ \delta \in A\), \(\beta \prec \gamma \in B\), and \(\delta\) and \(\gamma\) are 2-cells. Thus, we obtain \(d_\delta^{A,t} = d_\gamma^{B,t}\). Also, by Proposition \ref{proposition:cyclic-shifting families}, \(C_{n-1}^{X,t + n - 2}(\sigma) = C_{n-1}^{Y, t + n - 2}(\tau)\) implies \(C_1^{X,t}(\sigma) = C_1^{Y,t}(\tau)\), which is equivalent to \(d_\mathcal{B}^{A,t}(\delta) = d_\mathcal{B}^{B,t}(\gamma)\).

    For every 2-cell that a 1-cell is adjacent to, we concatenate them to form a tuple, then form a multi-set containing those tuples:
\[
    \left\{\!\!\left\{ (d_\varepsilon^{A,t}, d_\delta^{A,t})  \right\}\!\!\right\} = \left\{\!\!\left\{ (d_\zeta^{B,t}, d_\gamma^{B,t})  \right\}\!\!\right\}   
\]
where \(\varepsilon \in \mathcal{B}(\delta)\) and \(\zeta \in \mathcal{B}(\gamma)\).

    Merge all these multi-sets (because there can be many 2-cells that a 1-cell is adjacent to), we obtain:
    \[
        d_\uparrow^{A,t}(\alpha) = d_\uparrow^{B,t}(\beta)
  \]
  Finally, we obtain \(d_\alpha^{A,t+1} = d_\beta^{B, t+1}\) because \(d_\alpha^{A,t} = d_\beta^{B,t}\), \(d_{\mathcal{B}}^{A,t}(\alpha)=d_{\mathcal{B}}^{B,t}(\beta)\), and \(d_\uparrow^{A,t}(\alpha) = d_\uparrow^{B,t}(\beta)\).
    
\end{proof}
\subsection{Proof for Theorem \ref{theorem:pwl_cwl}}

\begin{proof}
    By Lemma \ref{lemma:p-paths to 0-cells}, \ref{lemma:p-paths to 2-cells}, and \ref{lemma:p-paths to 1-cells}, colorings of cells performed by CWL(\(k\text{-IC}\)) with maximum dimension of 2 can be determined by colorings of elementary paths performed by PWL.
\end{proof}

\subsection{Proof for Corollary \ref{corollary:pwl_wl}}

\begin{proof}
    PWL is provably as powerful as CWL(\(k\)-IC), and CWL is proven to be strictly more powerful than WL in \cite{bodnar_weisfeiler_2022}.
\end{proof}

\subsection{Proof for Corollary \ref{corollary:pwl_3wl}}

\begin{proof}
    Graphs from the same SRG family cannot be distinguished by the 3-WL test as proven in \cite{bodnar_weisfeiler_2021}; however, there exists a pair of graphs that can be distinguished by SWL and CWL \cite{bodnar_weisfeiler_2021, bodnar_weisfeiler_2022}. Thus, as PWL is at least as powerful as SWL and CWL(\(k\text{-IC}\)), PWL is not less powerful than 3-WL.
\end{proof}

\section{Formula for Higher-order Message-passing Neural Networks} \label{app:formula-mp}

The update formula of PIN, as well as SIN \cite{bodnar_weisfeiler_2021} and CIN \cite{bodnar_weisfeiler_2022}, is expressed as follow:

\begin{gather*}
    h_{\sigma}^{(t+1)} = \text{MLP}_{\text{UP},p}^{(t)} \left( m_{\mathcal{B}}^{(t)}(\sigma) \,\vert\vert\, m_{\uparrow}^{(t)}(\sigma) \right) \\
m_{\mathcal{B}}^{(t)}(\sigma) = \text{MLP}_{\mathcal{B},p}^{(t)}\left(\left(1 + \varepsilon_{\mathcal{B}}\right)h_{\sigma}^{(t)} + \sum_{\tau \in \mathcal{B}(\sigma)}h_{\tau}^{(t)} \right) \end{gather*}
\begin{equation*}
\begin{aligned}
m_{\uparrow}^{(t)}(\sigma) =\text{MLP}_{\mathcal{\uparrow},p}^{(t)}\Bigg(\left(1 + \varepsilon_{\mathcal{\uparrow}}\right)h_{\sigma}^{(t)}
+ \\
\sum_{\substack{\tau \in \mathcal{N}_{\uparrow}(\sigma) \\ \delta \in \mathcal{C}(\sigma,\tau)}}\text{MLP}_{M,p}^{(t)}\left(h_{\tau}^{(t)} \, \vert\vert \, h_{\delta}^{(t)}\right)\Bigg)
\end{aligned}
\end{equation*}
where \(\sigma\) is an elementary \(p\)-path (\(p\)-simplex for SIN \cite{bodnar_weisfeiler_2021} or \(p\)-cell for \cite{bodnar_weisfeiler_2022}). If the features from co-boundaries are not considered, we have an isotropic model in which messages from upper-adjacent neighborhoods are aggregated by summation. After propagating messages between elementary paths, we leverage permutation-invariant pooling function, which is summation or mean, followed by a single dense layer to get the representation of a certain dimension. To get the final representation of a path complex, we apply a \textsc{Readout} function on all representations of dimensions:
\[
    h_P = \textsc{Readout}\left(\left\{\!\left\{ h_{P_i} \, | \, i \leq n\right\}\!\right\}\right)
\]
where \textsc{Readout} can be sum, mean, or learnable weighted summation powered by a multi-layer perceptron \cite{giusti_cin_2023}. The final representation of a path complex is fed into a 2-layer or 3-layer multi-layer perceptron to get the output optimized by a certain loss function: 
\[
    h_G = \text{MLP}_{\text{proj}}\left( h_P \right)
\]
Additionally, we include several tricks to further boost the model's performance such as Jumping Knowledge\cite{xu_representation_2018} or Dropout \cite{srivastava_dropout_2014}, as suggested by \cite{bodnar_weisfeiler_2021, bodnar_weisfeiler_2022}.

\section{Experiments} \label{app:experiments}
Our code is based on existing works on MPSN \cite{bodnar_weisfeiler_2021} and CWN \cite{bodnar_weisfeiler_2022}, which are  powered by Pytorch \cite{paszke_pytorch_2019} and Pytorch Geometric \cite{fey_fast_2019}. The graph lifting transformation is implemented with the help of graph-tool library \cite{peixoto_graph-tool_2014} and NetworkX \cite{hagberg_exploring_2008}. Unless stated otherwise, all of our experiments are optimized by AdamW \cite{loschilov_decoupled_2019}. All of the experiments are executed on 4 machines: 1) Intel\textsuperscript{\textregistered} Xeon\textsuperscript{\textregistered} Silver 4214R CPU @ 2.40GHz and NVIDIA\textsuperscript{\textregistered} Quadro RTX\textsuperscript{TM} 8000, 2) Intel\textsuperscript{\textregistered} Xeon\textsuperscript{\textregistered} Silver 4114 CPU @ 2.20GHz and NVIDIA\textsuperscript{\textregistered} GeForce\textsuperscript{\textregistered} RTX\textsuperscript{TM} 2080 Ti, 3) Intel\textsuperscript{\textregistered} Core\textsuperscript{TM} i9-9820X CPU @ 3.30GHz and NVIDIA\textsuperscript{\textregistered} TITAN RTX\textsuperscript{TM}, and 4) Intel\textsuperscript{\textregistered} Xeon\textsuperscript{\textregistered} W-2145 CPU @ 3.70GHz and NVIDIA\textsuperscript{\textregistered} GeForce\textsuperscript{\textregistered} RTX\textsuperscript{TM} 2080 Ti.

\subsection{TUDataset Benchmarks} \label{subsec:appendix-tu}

Since we observe that our model overfits the PROTEIN dataset, we remove all non-linear activations from the message-passing layers, while still keeping non-linear activations for the last fully connected layers. Following the precedent set by \cite{bodnar_weisfeiler_2021,bodnar_weisfeiler_2022}, we populate higher-dimensional elementary \(p\)-paths in a recursive fashion, in which feature of the current elementary \(p\)-path is either the summation or mean of the elementary \((p-1)\)-paths on its boundary. All hyperparameter settings are fine-tuned with WandB \cite{wandb} by the random search method, and are reported in Table \ref{tab:tu-params}. Dropout \cite{srivastava_dropout_2014} is applied before the last fully connected layer if applicable.

\begin{table}
\centering
\resizebox{\columnwidth}{!}{%
\begin{tabular}{@{}lcccc@{}}
\toprule
Parameter                                                                          & PROTEINS                 & NCI1                     & NCI109                   & IMDB-B                   \\ \midrule
Batch size                                                                         & 128                      & 128                      & 128                      & 128                      \\
Dropout rate                                                                       & 0.0                        & 0.0                        & 0.0                        & 0.4                      \\
Embedding dimension                                                                & 16                       & 32                       & 64                       & 16                       \\
Final \textsc{Readout}                                                             & sum                      & sum                      & sum                      & sum                      \\
Initialization method                                                              & sum                      & sum                      & mean                     & sum                      \\
Jumping knowledge                                                                  & None                     & None                     & Concatenation            & None                     \\
Learning rate                                                                      & 0.005                    & 0.0005                   & 0.0005                   & 0.001                    \\
Learning decay rate                                                                & 0.6                      & 0.2                      & 0.4                      & 0.2                      \\
Learning decay step                                                                & 40                       & 60                       & 20                       & 50                       \\
Maximum lifting dimension                                                          & 2                        & 7                        & 3                        & 2                        \\
\# Layers in MLP\textsubscript{proj}                                                  & 3                        & 2                        & 2                        & 2                        \\
\# Message-passing layers                                                           & 3                        & 2                        & 7                        & 4                        \\
\# Layers in MLP\textsubscript{\(\uparrow\)} and MLP\textsubscript{\(\mathcal{B}\)} & 2                        & 2                        & 1                        & 1                        \\
\# Layers in MLP\textsubscript{UP}                                                  & 1                        & 1                        & 1                        & 1                        \\
Path-level \textsc{Readout}                                                        & sum                      & sum                      & sum                      & sum                      \\
Use co-boundaries                                                                  & FALSE                    & FALSE                    & FALSE                    & FALSE                    \\ \bottomrule
\end{tabular}%
}
\caption{Hyperparameter settings for TUDataset Benchmarks.}
\label{tab:tu-params}
\end{table}
\begin{table}[t]
\centering
\begin{tabular}{@{}lcc@{}}
\toprule
     & \multicolumn{1}{l}{PROTEINS} & \multicolumn{1}{l}{IMDB-B} \\ \midrule
Mean & 77.72                        & 75.37                      \\
Std  & 0.43                         & 0.57                       \\
Min  & 77.2                         & 74.7                       \\
Max  & 78.8                         & 76.6                       \\ \bottomrule
\end{tabular}%
\caption{Statistics of 10 runs of 10-fold experiments on PROTEINS and IMDB-B.}
\label{tab:proteins-imdbb}
\end{table}

\begin{table}[t]
\centering
\resizebox{\columnwidth}{!}{%
\begin{tabular}{@{}lcc@{}}
\toprule
Parameter                                                                           & ZINC                     & OGBG-MOLHIV                   \\ \midrule
Batch size                                                                          & 128                      & 128                      \\
Embedding dimension                                                                 & 64                       & 64                       \\
Final \textsc{Readout}                                                              & sum                      & sum                      \\
Initialization method                                                               & sum                      & sum                      \\
Jumping knowledge                                                                   & None                     & None                     \\
Learning rate scheduler                                                             & ReduceLROnPlateau        & None                     \\
Learning rate                                                                       & 0.001                    & 0.0001                   \\
Learning decay rate     & 0.5 & None \\
Learning decay patience                                                             & 20                       & None                     \\
Maximum lifting dimension                                                           & 7                        & 2                        \\
\# Layers in MLP\textsubscript{proj}                                                & 2                        & 2                        \\
\# Message-passing layers                                                           & 4                        & 1                        \\
\# Layers in MLP\textsubscript{\(\uparrow\)} and MLP\textsubscript{\(\mathcal{B}\)} & 2                        & 2                        \\
\# Layers in MLP\textsubscript{UP}                                                  & 1                        & 1                        \\
\# Layers in MLP\textsubscript{M} & 1 & 1 \\
Path-level \textsc{Readout}                                                         & sum                      & mean                     \\
Use co-boundaries                                                                   & TRUE                     & TRUE                     \\ \bottomrule
\end{tabular}%
}
\caption{Hyperparameter settings for ZINC and OGBG-MOLHIV.}
\label{tab:zinc-molhiv-params}
\end{table}

It is worth noting that PROTEINS and IMDB-B are small datasets with the total number of graphs is about 1000. Thus, we observe significant variations in the final results between 10-fold runs even with identical hyperparameter and randomness settings. Such behavior, which is due to floating point uncertainty in parallel computation engaged in message passing, is common in small datasets and well-documented by Pytorch \cite{paszke_pytorch_2019} and Pytorch Geometric \cite{fey_fast_2019}. Thus, we repeat the PROTEINS and IMDB-B experiments 10 more times, and report their statistics in Table \ref{tab:proteins-imdbb}. We do not observe significant variations in performance for other datasets.

\begin{table}[t]
\centering
\begin{tabular}{@{}ccc@{}}
\toprule
Dimension & Validation    & Test          \\ \midrule
2         & 0.172 ± 0.010 & 0.160 ± 0.009 \\
3         & 0.165 ± 0.007 & 0.151 ± 0.010 \\
4         & 0.164 ± 0.008 & 0.146 ± 0.008 \\
5         & 0.163 ± 0.010 & 0.146 ± 0.008 \\
6         & 0.146 ± 0.016 & 0.120 ± 0.010 \\
7         & 0.127 ± 0.009 & 0.096 ± 0.006 \\ \bottomrule
\end{tabular}
\caption{Performance on Validation and Test Sets of ZINC (with edge features) by varying the maximum lifting dimension.}
\label{tab:zinc-vary-dim}
\end{table}

\subsection{ZINC}

ZINC is introduced by \cite{sterling_zinc15_2015}, which contains up to 250,000 molecular graphs. \cite{dwivedi_benchmarkgnns_2020} suggests splitting the ZINC dataset into two benchmarks: ZINC-FULL, which contains all 250,000 molecular graphs, and ZINC, which is a subset of ZINC-FULL which contains 12,000 molecular graphs. On average, a graph may contain 2 or 3 rings, while the maximum number of rings that a graph may have is 21. Regarding ring size, a majority of rings have sizes of 5 or 6, and the max ring size is 18.

We lift graphs to a dimension of 7, and features of higher-dimensional elementary paths are computed in the same manner as established in the TUDataset Benchmarks. For this dataset, we use hyperparameters reported by \cite{bodnar_weisfeiler_2022}, with an exception that the embedding dimension is reduced to 64. Table \ref{tab:zinc-molhiv-params} reports detailed hyperparameter settings.
\begin{table}[t]
\centering
\resizebox{\columnwidth}{!}{%
\begin{tabular}{@{}ccccc@{}}
\toprule
\multirow{2}{*}{Seed} & \multicolumn{2}{c}{Run 1}     & \multicolumn{2}{c}{Run 2}       \\ \cmidrule(l){2-5} 
                      & Validation    & Test          & Validation     & Test           \\ \midrule
0                     & 0.102         & 0.086         & 0.130          & 0.106          \\
1                     & 0.119         & 0.097         & 0.127          & 0.109          \\
2                     & 0.128         & 0.096         & 0.110          & 0.079          \\
3                     & 0.116         & 0.091         & 0.119          & 0.084          \\
4                     & 0.126         & 0.100         & 0.133          & 0.102          \\
5                     & 0.119         & 0.085         & 0.109          & 0.087          \\
6                     & 0.115         & 0.089         & 0.118          & 0.089          \\
7                     & 0.109         & 0.080         & 0.116          & 0.094          \\
8                     & 0.104         & 0.090         & \textbf{0.176} & \textbf{0.153} \\
9                     & 0.105         & 0.088         & 0.109          & 0.092          \\ \midrule
MAE          & 0.114 ± 0.009 & 0.090 ± 0.006 & 0.125 ± 0.020  & 0.100 ± 0.021  \\ \bottomrule
\end{tabular}%
}
\caption{Two different runs on the ZINC dataset with the number of message-passing layers of 6. Bold text indicates the performance that deviates from the rest.}
\label{tab:zinc-uncertainty}
\end{table}
We also evaluate the effectiveness of higher orders on performance. Table \ref{tab:zinc-vary-dim} demonstrates the monotonic increase in performance on the validation and test sets of the ZINC dataset, while the number of message-passing layers is fixed to 4.

In fact, we are able to obtain a better performance than the one reported in Table \ref{tab:zinc-molhiv} by increasing the number of message-passing layers to 6. However, we observe that model occasionally has poor performance that greatly deviates from others.  We hypothesize that such an uncertainty is introduced by the parallel operation as stated in Section \ref{subsec:appendix-tu}, as populating higher-order embeddings invoke this operation substantially and stacking more message-passing layers worsen the problem. However, to give a definite answer to this behavior requires an in-depth investigation. We do not observe such behavior when the number of message-passing layers is 4 or smaller. Thus, we only report in Table \ref{tab:zinc-molhiv} the performance when the number of message-passing layers is 4, where the uncertainty in performance does not exist. Table \ref{tab:zinc-uncertainty} demonstrates two different runs on the ZINC dataset with identical hyperparameter and randomness settings when the number of message-passing layers is 6.

\subsection{OGBG-MOLHIV}

In this experiment, we lift graphs to a dimension of 2, and features of higher-dimensional elementary paths are populated similarly as in the TUDataset Benchmarks and ZINC. Table \ref{tab:zinc-molhiv-params} reports detailed hyperparameter settings.

\subsection{Strongly Regular Graphs}

For the SRG experiment, we compute the Euclidean distance between final representations of graphs after feeding them through PIN, and if the distance is below a pre-defined threshold \(\varepsilon\), we conclude that the model cannot determine if two graphs are non-isomorphic. In order to mimic the behavior of the PWL test, we leverage PIN with randomly initialized weights. Then, we feed-forward SRGs without back-propagation and compare the distance between embeddings of those graphs. Similar to \cite{bodnar_weisfeiler_2021, bodnar_weisfeiler_2022, bouritsas_improving_2021}, we consider a pair of graphs indistinguishable if their distance is below \(\varepsilon = 0.01\). We follow the same protocol outlined in \cite{bodnar_weisfeiler_2022}; the only difference is that our final embedding vector has a length of 32. A thing to note in this experiment is the time required to lift graphs in SR(35,16,6,8) is substantially expensive. Table \ref{tab:details-sr-a} and Table \ref{tab:details-sr-b} document detailed statistics for the SRG experiments demonstrated in Figure \ref{fig:sr-graphs}.

All of the above experiments are conducted on the workstation with Intel\textsuperscript{\textregistered} Core\textsuperscript{TM} i9-9820X CPU @ 3.30GHz and NVIDIA\textsuperscript{\textregistered} TITAN RTX\textsuperscript{TM}. The experiments share the same architecture, in which the batch size is 8, the hidden embeddings have dimension of 16, the number of layers in \(\text{MLP}_{\text{UP}}\) is 1, the number of layers in \(\text{MLP}_{\mathcal{B}}\) and \(\text{MLP}_{\uparrow}\) is 1, the number of layers in \(\text{MLP}_{\text{M}}\) is 1, the number of layers in \(\text{MLP}_\text{proj}\) is 2, and the non-linearity is ELU \cite{clevert_fast_2016}. Higher-order features are populated by summation of features on the boundary.

\section{Computational Analysis}

\begin{table*}[t]
\centering
\begin{tabular}{@{}lcccccc@{}}
\toprule
\multirow{2}{*}{Model} & \multicolumn{3}{c}{ZINC}                    & \multirow{2}{*}{NCI1} & \multirow{2}{*}{NCI109} & \multirow{2}{*}{OGBG-MOLHIV} \\ \cmidrule(lr){2-4}
                       & Training & Validation & Test    &                       &                         &                         \\ \midrule
CIN(6-IC)              & 8.28 ± 0.26  & 1.14 ± 0.17    & 1.08 ± 0.21 & 3.55 ± 0.12           & 3.59 ± 0.16             & 32.98 ± 1.02            \\
PIN(2)                 & 8.19 ± 0.29  & 1.08 ± 0.22    & 1.15 ± 0.24 & 3.55 ± 0.11           & 3.58 ± 0.19             & 33.51 ± 0.95            \\
PIN(4)                 & 12.66 ± 0.43 & 1.47 ± 0.31    & 1.46 ± 0.32 & 6.34 ± 0.17           & 6.45 ± 0.16             & 54.57 ± 1.84            \\
PIN(6)                 & 18.95 ± 0.48 & 2.04 ± 0.37    & 2.07 ± 0.37 & 12.58 ± 0.29          & 12.88 ± 0.32            & 95.36 ± 1.18            \\ \bottomrule
\end{tabular}%
\caption{Time (in seconds) to lift graphs.}
\label{tab:lift-graph}
\vspace{5mm}

\begin{tabular}{@{}lcccccccc@{}}
\toprule
\multirow{2}{*}{Model} & \multicolumn{2}{c}{ZINC}                             & \multicolumn{2}{c}{NCI1}                             & \multicolumn{2}{c}{NCI109}                       & \multicolumn{2}{c}{OGBG-MOLHIV}                       \\ \cmidrule(l){2-9} 
                       & Training & Inference  & Training & Inference & Training& Inference & Training & Inference \\ \midrule
CIN(6-IC)              & 3.46 ± 0.08             & 0.17 ± 0.01                & 1.12 ± 0.06             & 0.08 ± 0.01                & 1.15 ± 0.07             & 0.07 ± 0.01            & 12.26 ± 0.20            & 0.71 ± 0.01            \\
PIN(2)                 & 3.44 ± 0.10             & 0.17 ± 0.01                & 1.18 ± 0.07             & 0.07 ± 0.01                & 1.25 ± 0.05             & 0.07 ± 0.01            & 13.12 ± 0.21            & 0.77 ± 0.01            \\
PIN(4)                 & 6.30 ± 0.20             & 0.34 ± 0.02                & 2.10 ± 0.10             & 0.13 ± 0.00                & 2.12 ± 0.06             & 0.14 ± 0.01            & 22.11 ± 0.45            & 1.49 ± 0.02            \\
PIN(6)                 & 9.09 ± 0.20             & 0.51 ± 0.01                & 2.99 ± 0.10             & 0.20 ± 0.01                & 3.07 ± 0.08             & 0.21 ± 0.03            & 31.89 ± 0.30 & 2.26 ± 0.02    \\ \bottomrule
\end{tabular}%
\caption{Time (in seconds) to train 1 epoch and perform inference on validation sets.}
\label{tab:train-inference}
\vspace{5mm}
\centering
\begin{tabular}{@{}ccc@{}}
\toprule
Dimension              & NCI1 (Accuracy) & ZINC (MAE)    \\ \midrule
7                      & 85.1 ± 1.5      & 0.096 ± 0.006    \\
3                      & 84.9 ± 1.6      & 0.126 ± 0.004 \\ \midrule
Performance Difference & -0.2            & +0.03         \\ \bottomrule
\end{tabular}%
\caption{Performance Trade-offs on test sets when lifting to a lower dimension.}
\label{tab:nci1-zinc}
\end{table*}

From a theoretical perspective, the task of enumerating all paths of length \(k\) in a graph comprising \(n\) nodes presents a worst-case time complexity of \(\mathcal{O}(n^{k+1})\) or \(\mathcal{O}(nb^k)\) if taking the branching factor \(b\) of the graph into account. However, we found that NCI1, NCI109, ZINC, and OGBG-MOLHIV can be lifted to higher dimensions without incurring significant computational overhead. In this experiment, we lift graphs 10 times, and then compute the statistics of each lifting transformation. Table \ref{tab:lift-graph} reports total time required to lift graphs to a certain dimension for each of the above datasets. It is worth noting that the graph lifting operation is invoked during the first run only, and constructed path complexes are stored on a local storage for future runs.

In relation to the message-passing operation, as number of paths increases exponentially, time required for message aggregation also increases. While an efficient parallel computing implementation for higher-order message-passing are proposed by \cite{bodnar_weisfeiler_2021, bodnar_weisfeiler_2022} using tools such as Pytorch Geometric \cite{fey_fast_2019} and Pytorch \cite{paszke_pytorch_2019}, message-passing still depends on the number of elementary paths for each dimension. Similar to the above experiment, we compute the mean elapsed time per epoch, while for inference, we calculate the average time taken for inference on validation sets, over 10 runs. Table \ref{tab:train-inference} reports time taken for each iteration of training and inference on different datasets.

All of the above experiments are conducted on the workstation with Intel\textsuperscript{\textregistered} Core\textsuperscript{TM} i9-9820X CPU @ 3.30GHz and NVIDIA\textsuperscript{\textregistered} TITAN RTX\textsuperscript{TM}. The experiments share the same architecture, in which the batch size is 128, the hidden embeddings have dimension of 32, the number of message-passing layers is 2, the number of layers in \(\text{MLP}_{\text{UP}}\) is 1, the number of layers in \(\text{MLP}_{\mathcal{B}}\) and \(\text{MLP}_{\uparrow}\) is 2, the number of layers in \(\text{MLP}_{\text{M}}\) is 1 (applicable for ZINC and OGBG-MOLHIV only), and the number of layers in \(\text{MLP}_\text{proj}\) is 2.

It is worth noting that lifting to higher dimensions should not be strictly adhered to, as we can still obtain a robust higher-order representation even with low dimensions. We define any dimension less than 4 low, as it will not incur a major burden on computation as illustrated by Table \ref{tab:lift-graph} and Table \ref{tab:train-inference}. As shown by Table \ref{tab:tu-params} and Table \ref{tab:zinc-molhiv-params}, the maximum dimension graphs lifted to are 2 for PROTEINS, IMDB-B, and OGBG-MOLHIV and 3 for NCI109. For ZINC and NCI1, we found that lifting graphs to a dimension of 3 is sufficient to outperform other state-of-the-art methods shown in Table \ref{tab:tu-datasets} and Table \ref{tab:zinc-molhiv}. Table \ref{tab:nci1-zinc} demonstrates the trade-offs. In this experiment, for the ZINC dataset, our model has the embedding dimension of 128, the number of message-passing layers is 6, and other hyperaparameters remain the same. For the NCI1 dataset, only the maximum lifting dimension is changed.

\section{Other Resources}
We leverage GPT-4 by OpenAI \cite{gpt4_2022} as an assistive writing tool in the development of this paper. We use WandB \cite{wandb} to keep track of research experiments and perform hyperparameter searches.

\begin{table*}[t]
\centering
\resizebox{\textwidth}{!}{%
\begin{tabular}{@{}ccccc|ccccc@{}}
\toprule
\multicolumn{2}{c}{\multirow{2}{*}{\textbf{3 Layers}}}                          & \multicolumn{3}{c|}{Model}       & \multicolumn{2}{c}{\multirow{2}{*}{\textbf{4 Layers}}}                          & \multicolumn{3}{c}{Model}        \\ \cmidrule(lr){3-5} \cmidrule(l){8-10} 
\multicolumn{2}{c}{}                                                            & CWN(4-IC) & CWN(5-IC) & PCN(3)   & \multicolumn{2}{c}{}                                                            & CWN(4-IC) & CWN(5-IC) & PCN(3)   \\ \midrule
\multicolumn{1}{c|}{\multirow{4}{*}{SR(16,6,2,2)}}  & \multicolumn{1}{c|}{Mean} & 0         & 0         & 0        & \multicolumn{1}{c|}{\multirow{4}{*}{SR(16,6,2,2)}}  & \multicolumn{1}{c|}{Mean} & 0         & 0         & 0        \\
\multicolumn{1}{c|}{}                               & \multicolumn{1}{c|}{Std}  & 0         & 0         & 0        & \multicolumn{1}{c|}{}                               & \multicolumn{1}{c|}{Std}  & 0         & 0         & 0        \\
\multicolumn{1}{c|}{}                               & \multicolumn{1}{c|}{Min}  & 0         & 0         & 0        & \multicolumn{1}{c|}{}                               & \multicolumn{1}{c|}{Min}  & 0         & 0         & 0        \\
\multicolumn{1}{c|}{}                               & \multicolumn{1}{c|}{Max}  & 0         & 0         & 0        & \multicolumn{1}{c|}{}                               & \multicolumn{1}{c|}{Max}  & 0         & 0         & 0        \\ \midrule
\multicolumn{1}{c|}{\multirow{4}{*}{SR(25,12,5,6)}} & \multicolumn{1}{c|}{Mean} & 0         & 0         & 0.00286  & \multicolumn{1}{c|}{\multirow{4}{*}{SR(25,12,5,6)}} & \multicolumn{1}{c|}{Mean} & 0         & 0         & 0        \\
\multicolumn{1}{c|}{}                               & \multicolumn{1}{c|}{Std}  & 0         & 0         & 0.00086  & \multicolumn{1}{c|}{}                               & \multicolumn{1}{c|}{Std}  & 0         & 0         & 0        \\
\multicolumn{1}{c|}{}                               & \multicolumn{1}{c|}{Min}  & 0         & 0         & 0        & \multicolumn{1}{c|}{}                               & \multicolumn{1}{c|}{Min}  & 0         & 0         & 0        \\
\multicolumn{1}{c|}{}                               & \multicolumn{1}{c|}{Max}  & 0         & 0         & 0.02857  & \multicolumn{1}{c|}{}                               & \multicolumn{1}{c|}{Max}  & 0         & 0         & 0        \\ \midrule
\multicolumn{1}{c|}{\multirow{4}{*}{SR(26,10,3,4)}} & \multicolumn{1}{c|}{Mean} & 0.04444   & 0         & 0.02444  & \multicolumn{1}{c|}{\multirow{4}{*}{SR(26,10,3,4)}} & \multicolumn{1}{c|}{Mean} & 0.04222   & 0         & 0.00667  \\
\multicolumn{1}{c|}{}                               & \multicolumn{1}{c|}{Std}  & 0         & 0         & 0.0012   & \multicolumn{1}{c|}{}                               & \multicolumn{1}{c|}{Std}  & 0.00067   & 0         & 0.00102  \\
\multicolumn{1}{c|}{}                               & \multicolumn{1}{c|}{Min}  & 0.04444   & 0         & 0        & \multicolumn{1}{c|}{}                               & \multicolumn{1}{c|}{Min}  & 0.02222   & 0         & 0        \\
\multicolumn{1}{c|}{}                               & \multicolumn{1}{c|}{Max}  & 0.04444   & 0         & 0.04444  & \multicolumn{1}{c|}{}                               & \multicolumn{1}{c|}{Max}  & 0.04444   & 0         & 0.02222  \\ \midrule
\multicolumn{1}{c|}{\multirow{4}{*}{SR(28,12,6,4)}} & \multicolumn{1}{c|}{Mean} & 0         & 0         & 0        & \multicolumn{1}{c|}{\multirow{4}{*}{SR(28,12,6,4)}} & \multicolumn{1}{c|}{Mean} & 0         & 0         & 0        \\
\multicolumn{1}{c|}{}                               & \multicolumn{1}{c|}{Std}  & 0         & 0         & 0        & \multicolumn{1}{c|}{}                               & \multicolumn{1}{c|}{Std}  & 0         & 0         & 0        \\
\multicolumn{1}{c|}{}                               & \multicolumn{1}{c|}{Min}  & 0         & 0         & 0        & \multicolumn{1}{c|}{}                               & \multicolumn{1}{c|}{Min}  & 0         & 0         & 0        \\
\multicolumn{1}{c|}{}                               & \multicolumn{1}{c|}{Max}  & 0         & 0         & 0        & \multicolumn{1}{c|}{}                               & \multicolumn{1}{c|}{Max}  & 0         & 0         & 0        \\ \midrule
\multicolumn{1}{c|}{\multirow{4}{*}{SR(29,14,6,7)}} & \multicolumn{1}{c|}{Mean} & 0         & 0         & 0.00037  & \multicolumn{1}{c|}{\multirow{4}{*}{SR(29,14,6,7)}} & \multicolumn{1}{c|}{Mean} & 0         & 0         & 0        \\
\multicolumn{1}{c|}{}                               & \multicolumn{1}{c|}{Std}  & 0         & 0         & 0.00011  & \multicolumn{1}{c|}{}                               & \multicolumn{1}{c|}{Std}  & 0         & 0         & 0        \\
\multicolumn{1}{c|}{}                               & \multicolumn{1}{c|}{Min}  & 0         & 0         & 0        & \multicolumn{1}{c|}{}                               & \multicolumn{1}{c|}{Min}  & 0         & 0         & 0        \\
\multicolumn{1}{c|}{}                               & \multicolumn{1}{c|}{Max}  & 0         & 0         & 0.00366  & \multicolumn{1}{c|}{}                               & \multicolumn{1}{c|}{Max}  & 0         & 0         & 0        \\ \midrule
\multicolumn{1}{c|}{\multirow{4}{*}{SR(35,16,6,8)}} & \multicolumn{1}{c|}{Mean} & 1.55E-06  & 0         & 1.07E-05 & \multicolumn{1}{c|}{\multirow{4}{*}{SR(35,16,6,8)}} & \multicolumn{1}{c|}{Mean} & 5.39E-08  & 0         & 1.71E-06 \\
\multicolumn{1}{c|}{}                               & \multicolumn{1}{c|}{Std}  & 1.69E-07  & 0         & 5.74E-07 & \multicolumn{1}{c|}{}                               & \multicolumn{1}{c|}{Std}  & 6.60E-09  & 0         & 2.97E-07 \\
\multicolumn{1}{c|}{}                               & \multicolumn{1}{c|}{Min}  & 4.04E-07  & 0         & 3.37E-06 & \multicolumn{1}{c|}{}                               & \multicolumn{1}{c|}{Min}  & 0         & 0         & 0        \\
\multicolumn{1}{c|}{}                               & \multicolumn{1}{c|}{Max}  & 6.46E-06  & 0         & 1.94E-05 & \multicolumn{1}{c|}{}                               & \multicolumn{1}{c|}{Max}  & 1.35E-07  & 0         & 1.05E-05 \\ \midrule
\multicolumn{1}{c|}{\multirow{4}{*}{SR(35,18,9,9)}} & \multicolumn{1}{c|}{Mean} & 0         & 0         & 6.24E-05 & \multicolumn{1}{c|}{\multirow{4}{*}{SR(35,18,9,9)}} & \multicolumn{1}{c|}{Mean} & 0         & 0         & 0        \\
\multicolumn{1}{c|}{}                               & \multicolumn{1}{c|}{Std}  & 0         & 0         & 3.98E-06 & \multicolumn{1}{c|}{}                               & \multicolumn{1}{c|}{Std}  & 0         & 0         & 0        \\
\multicolumn{1}{c|}{}                               & \multicolumn{1}{c|}{Min}  & 0         & 0         & 0        & \multicolumn{1}{c|}{}                               & \multicolumn{1}{c|}{Min}  & 0         & 0         & 0        \\
\multicolumn{1}{c|}{}                               & \multicolumn{1}{c|}{Max}  & 0         & 0         & 0.00012  & \multicolumn{1}{c|}{}                               & \multicolumn{1}{c|}{Max}  & 0         & 0         & 0        \\ \midrule
\multicolumn{1}{c|}{\multirow{4}{*}{SR(36,14,4,6)}} & \multicolumn{1}{c|}{Mean} & 0.01906   & 0         & 0.00096  & \multicolumn{1}{c|}{\multirow{4}{*}{SR(36,14,4,6)}} & \multicolumn{1}{c|}{Mean} & 0.01147   & 0         & 0.0002   \\
\multicolumn{1}{c|}{}                               & \multicolumn{1}{c|}{Std}  & 2.83E-05  & 0         & 6.12E-05 & \multicolumn{1}{c|}{}                               & \multicolumn{1}{c|}{Std}  & 0.00025   & 0         & 2.91E-05 \\
\multicolumn{1}{c|}{}                               & \multicolumn{1}{c|}{Min}  & 0.01844   & 0         & 6.21E-05 & \multicolumn{1}{c|}{}                               & \multicolumn{1}{c|}{Min}  & 0.00931   & 0         & 0        \\
\multicolumn{1}{c|}{}                               & \multicolumn{1}{c|}{Max}  & 0.01962   & 0         & 0.00217  & \multicolumn{1}{c|}{}                               & \multicolumn{1}{c|}{Max}  & 0.01887   & 0         & 0.00099  \\ \midrule
\multicolumn{1}{c|}{\multirow{4}{*}{SR(40,12,2,4)}} & \multicolumn{1}{c|}{Mean} & 0.01642   & 0.00529   & 0.02407  & \multicolumn{1}{c|}{\multirow{4}{*}{SR(40,12,2,4)}} & \multicolumn{1}{c|}{Mean} & 0.01085   & 0.00529   & 0.00688  \\
\multicolumn{1}{c|}{}                               & \multicolumn{1}{c|}{Std}  & 0.00023   & 0         & 0.00211  & \multicolumn{1}{c|}{}                               & \multicolumn{1}{c|}{Std}  & 0.00028   & 0         & 0.00021  \\
\multicolumn{1}{c|}{}                               & \multicolumn{1}{c|}{Min}  & 0.01058   & 0.00529   & 0.00529  & \multicolumn{1}{c|}{}                               & \multicolumn{1}{c|}{Min}  & 0.00794   & 0.00529   & 0.00529  \\
\multicolumn{1}{c|}{}                               & \multicolumn{1}{c|}{Max}  & 0.01852   & 0.00529   & 0.08201  & \multicolumn{1}{c|}{}                               & \multicolumn{1}{c|}{Max}  & 0.01587   & 0.00529   & 0.01058  \\ \bottomrule
\end{tabular}%
}
\caption{Detailed statistics of the SRG experiments (3 message-passing layers and 4 message-passing layers)}
\label{tab:details-sr-a}
\end{table*}
\begin{table*}[t]
\centering
\resizebox{\textwidth}{!}{%
\begin{tabular}{@{}ccccc|ccccc@{}}
\toprule
\multicolumn{2}{c}{\multirow{2}{*}{\textbf{5 Layers}}}                          & \multicolumn{3}{c|}{Model}       & \multicolumn{2}{c}{\multirow{2}{*}{\textbf{6 Layers}}}                          & \multicolumn{3}{c}{Model}        \\ \cmidrule(lr){3-5} \cmidrule(l){8-10} 
\multicolumn{2}{c}{}                                                            & CWN(4-IC) & CWN(5-IC) & PCN(3)   & \multicolumn{2}{c}{}                                                            & CWN(4-IC) & CWN(5-IC) & PCN(3)   \\ \midrule
\multicolumn{1}{c|}{\multirow{4}{*}{SR(16,6,2,2)}}  & \multicolumn{1}{c|}{Mean} & 0         & 0         & 0        & \multicolumn{1}{c|}{\multirow{4}{*}{SR(16,6,2,2)}}  & \multicolumn{1}{c|}{Mean} & 0         & 0         & 0        \\
\multicolumn{1}{c|}{}                               & \multicolumn{1}{c|}{Std}  & 0         & 0         & 0        & \multicolumn{1}{c|}{}                               & \multicolumn{1}{c|}{Std}  & 0         & 0         & 0        \\
\multicolumn{1}{c|}{}                               & \multicolumn{1}{c|}{Min}  & 0         & 0         & 0        & \multicolumn{1}{c|}{}                               & \multicolumn{1}{c|}{Min}  & 0         & 0         & 0        \\
\multicolumn{1}{c|}{}                               & \multicolumn{1}{c|}{Max}  & 0         & 0         & 0        & \multicolumn{1}{c|}{}                               & \multicolumn{1}{c|}{Max}  & 0         & 0         & 0        \\ \midrule
\multicolumn{1}{c|}{\multirow{4}{*}{SR(25,12,5,6)}} & \multicolumn{1}{c|}{Mean} & 0         & 0         & 0        & \multicolumn{1}{c|}{\multirow{4}{*}{SR(25,12,5,6)}} & \multicolumn{1}{c|}{Mean} & 0         & 0         & 0        \\
\multicolumn{1}{c|}{}                               & \multicolumn{1}{c|}{Std}  & 0         & 0         & 0        & \multicolumn{1}{c|}{}                               & \multicolumn{1}{c|}{Std}  & 0         & 0         & 0        \\
\multicolumn{1}{c|}{}                               & \multicolumn{1}{c|}{Min}  & 0         & 0         & 0        & \multicolumn{1}{c|}{}                               & \multicolumn{1}{c|}{Min}  & 0         & 0         & 0        \\
\multicolumn{1}{c|}{}                               & \multicolumn{1}{c|}{Max}  & 0         & 0         & 0        & \multicolumn{1}{c|}{}                               & \multicolumn{1}{c|}{Max}  & 0         & 0         & 0        \\ \midrule
\multicolumn{1}{c|}{\multirow{4}{*}{SR(26,10,3,4)}} & \multicolumn{1}{c|}{Mean} & 0.02667   & 0         & 0        & \multicolumn{1}{c|}{\multirow{4}{*}{SR(26,10,3,4)}} & \multicolumn{1}{c|}{Mean} & 0.02222   & 0         & 0        \\
\multicolumn{1}{c|}{}                               & \multicolumn{1}{c|}{Std}  & 0.00089   & 0         & 0        & \multicolumn{1}{c|}{}                               & \multicolumn{1}{c|}{Std}  & 0         & 0         & 0        \\
\multicolumn{1}{c|}{}                               & \multicolumn{1}{c|}{Min}  & 0.02222   & 0         & 0        & \multicolumn{1}{c|}{}                               & \multicolumn{1}{c|}{Min}  & 0.02222   & 0         & 0        \\
\multicolumn{1}{c|}{}                               & \multicolumn{1}{c|}{Max}  & 0.04444   & 0         & 0        & \multicolumn{1}{c|}{}                               & \multicolumn{1}{c|}{Max}  & 0.02222   & 0         & 0        \\ \midrule
\multicolumn{1}{c|}{\multirow{4}{*}{SR(28,12,6,4)}} & \multicolumn{1}{c|}{Mean} & 0         & 0         & 0        & \multicolumn{1}{c|}{\multirow{4}{*}{SR(28,12,6,4)}} & \multicolumn{1}{c|}{Mean} & 0         & 0         & 0        \\
\multicolumn{1}{c|}{}                               & \multicolumn{1}{c|}{Std}  & 0         & 0         & 0        & \multicolumn{1}{c|}{}                               & \multicolumn{1}{c|}{Std}  & 0         & 0         & 0        \\
\multicolumn{1}{c|}{}                               & \multicolumn{1}{c|}{Min}  & 0         & 0         & 0        & \multicolumn{1}{c|}{}                               & \multicolumn{1}{c|}{Min}  & 0         & 0         & 0        \\
\multicolumn{1}{c|}{}                               & \multicolumn{1}{c|}{Max}  & 0         & 0         & 0        & \multicolumn{1}{c|}{}                               & \multicolumn{1}{c|}{Max}  & 0         & 0         & 0        \\ \midrule
\multicolumn{1}{c|}{\multirow{4}{*}{SR(29,14,6,7)}} & \multicolumn{1}{c|}{Mean} & 0         & 0         & 0        & \multicolumn{1}{c|}{\multirow{4}{*}{SR(29,14,6,7)}} & \multicolumn{1}{c|}{Mean} & 0         & 0         & 0        \\
\multicolumn{1}{c|}{}                               & \multicolumn{1}{c|}{Std}  & 0         & 0         & 0        & \multicolumn{1}{c|}{}                               & \multicolumn{1}{c|}{Std}  & 0         & 0         & 0        \\
\multicolumn{1}{c|}{}                               & \multicolumn{1}{c|}{Min}  & 0         & 0         & 0        & \multicolumn{1}{c|}{}                               & \multicolumn{1}{c|}{Min}  & 0         & 0         & 0        \\
\multicolumn{1}{c|}{}                               & \multicolumn{1}{c|}{Max}  & 0         & 0         & 0        & \multicolumn{1}{c|}{}                               & \multicolumn{1}{c|}{Max}  & 0         & 0         & 0        \\ \midrule
\multicolumn{1}{c|}{\multirow{4}{*}{SR(35,16,6,8)}} & \multicolumn{1}{c|}{Mean} & 0         & 0         & 5.39E-08 & \multicolumn{1}{c|}{\multirow{4}{*}{SR(35,16,6,8)}} & \multicolumn{1}{c|}{Mean} & 0         & 0         & 0        \\
\multicolumn{1}{c|}{}                               & \multicolumn{1}{c|}{Std}  & 0         & 0         & 8.93E-09 & \multicolumn{1}{c|}{}                               & \multicolumn{1}{c|}{Std}  & 0         & 0         & 0        \\
\multicolumn{1}{c|}{}                               & \multicolumn{1}{c|}{Min}  & 0         & 0         & 0        & \multicolumn{1}{c|}{}                               & \multicolumn{1}{c|}{Min}  & 0         & 0         & 0        \\
\multicolumn{1}{c|}{}                               & \multicolumn{1}{c|}{Max}  & 0         & 0         & 2.69E-07 & \multicolumn{1}{c|}{}                               & \multicolumn{1}{c|}{Max}  & 0         & 0         & 0        \\ \midrule
\multicolumn{1}{c|}{\multirow{4}{*}{SR(35,18,9,9)}} & \multicolumn{1}{c|}{Mean} & 0         & 0         & 0        & \multicolumn{1}{c|}{\multirow{4}{*}{SR(35,18,9,9)}} & \multicolumn{1}{c|}{Mean} & 0         & 0         & 0        \\
\multicolumn{1}{c|}{}                               & \multicolumn{1}{c|}{Std}  & 0         & 0         & 0        & \multicolumn{1}{c|}{}                               & \multicolumn{1}{c|}{Std}  & 0         & 0         & 0        \\
\multicolumn{1}{c|}{}                               & \multicolumn{1}{c|}{Min}  & 0         & 0         & 0        & \multicolumn{1}{c|}{}                               & \multicolumn{1}{c|}{Min}  & 0         & 0         & 0        \\
\multicolumn{1}{c|}{}                               & \multicolumn{1}{c|}{Max}  & 0         & 0         & 0        & \multicolumn{1}{c|}{}                               & \multicolumn{1}{c|}{Max}  & 0         & 0         & 0        \\ \midrule
\multicolumn{1}{c|}{\multirow{4}{*}{SR(36,14,4,6)}} & \multicolumn{1}{c|}{Mean} & 0.00943   & 0         & 4.97E-05 & \multicolumn{1}{c|}{\multirow{4}{*}{SR(36,14,4,6)}} & \multicolumn{1}{c|}{Mean} & 0.00865   & 0         & 3.10E-05 \\
\multicolumn{1}{c|}{}                               & \multicolumn{1}{c|}{Std}  & 7.80E-05  & 0         & 7.75E-06 & \multicolumn{1}{c|}{}                               & \multicolumn{1}{c|}{Std}  & 4.07E-05  & 0         & 3.10E-06 \\
\multicolumn{1}{c|}{}                               & \multicolumn{1}{c|}{Min}  & 0.00801   & 0         & 0        & \multicolumn{1}{c|}{}                               & \multicolumn{1}{c|}{Min}  & 0.00788   & 0         & 0        \\
\multicolumn{1}{c|}{}                               & \multicolumn{1}{c|}{Max}  & 0.01105   & 0         & 0.00025  & \multicolumn{1}{c|}{}                               & \multicolumn{1}{c|}{Max}  & 0.00906   & 0         & 6.21E-05 \\ \midrule
\multicolumn{1}{c|}{\multirow{4}{*}{SR(40,12,2,4)}} & \multicolumn{1}{c|}{Mean} & 0.01085   & 0.00529   & 0.00556  & \multicolumn{1}{c|}{\multirow{4}{*}{SR(40,12,2,4)}} & \multicolumn{1}{c|}{Mean} & 0.00926   & 0.00529   & 0.00503  \\
\multicolumn{1}{c|}{}                               & \multicolumn{1}{c|}{Std}  & 0.00019   & 0         & 7.94E-05 & \multicolumn{1}{c|}{}                               & \multicolumn{1}{c|}{Std}  & 0.00013   & 0         & 7.94E-05 \\
\multicolumn{1}{c|}{}                               & \multicolumn{1}{c|}{Min}  & 0.00794   & 0.00529   & 0.00529  & \multicolumn{1}{c|}{}                               & \multicolumn{1}{c|}{Min}  & 0.00794   & 0.00529   & 0.00265  \\
\multicolumn{1}{c|}{}                               & \multicolumn{1}{c|}{Max}  & 0.01587   & 0.00529   & 0.00794  & \multicolumn{1}{c|}{}                               & \multicolumn{1}{c|}{Max}  & 0.01058   & 0.00529   & 0.00529  \\ \bottomrule
\end{tabular}%
}
\caption{Detailed statistics of the SRG experiments (5 message-passing layers and 6 message-passing layers)}
\label{tab:details-sr-b}
\end{table*}
\end{document}